# Contrastive Representation Learning: A Framework and Review

PHUC H. LE-KHAC[1], GRAHAM HEALY[2], AND ALAN F. SMEATON[2,3], (Fellow, IEEE)
[1]ML-Labs, Dublin City University, Dublin 9, D09 Ireland
[2]School of Computing, Dublin City University, Dublin 9, D09 Ireland
[3]Insight Centre for Data Analytics, Dublin City University, Dublin 9, D09 Ireland

Corresponding author: Phuc H. Le-Khac (khac.le2@mail.dcu.ie)

This work was supported in part by the Science Foundation Ireland through the Science Foundation Ireland (SFI) Centre for Research Training in Machine Learning (18/CRT/6183) and in part by the Insight Centre for Data Analytics (SFI/12/RC/2289_P2).

**ABSTRACT** Contrastive Learning has recently received interest due to its success in self-supervised representation learning in the computer vision domain. However, the origins of Contrastive Learning date as far back as the 1990s and its development has spanned across many fields and domains including Metric Learning and natural language processing. In this paper, we provide a comprehensive literature review and we propose a general Contrastive Representation Learning framework that simplifies and unifies many different contrastive learning methods. We also provide a taxonomy for each of the components of contrastive learning in order to summarise it and distinguish it from other forms of machine learning. We then discuss the inductive biases which are present in any contrastive learning system and we analyse our framework under different views from various sub-fields of Machine Learning. Examples of how contrastive learning has been applied in computer vision, natural language processing, audio processing, and others, as well as in Reinforcement Learning are also presented. Finally, we discuss the challenges and some of the most promising future research directions ahead.

## I. INTRODUCTION

The performance of a machine learning system is directly determined by the choice and quality of the data representation, or features, in the data used to train it. While it is obvious that some criteria for usefulness depend on the task, it is also universally assumed that there are sets of features that are representative of a dataset and that are generally useful as input for many kinds of downstream classifier or predictor. Focusing explicitly on learning representation in some cases can be beneficial, for example, when a labelled dataset for a task is small and we want to leverage a larger unlabelled dataset to improve the performance of a learning system.

*Representation learning* refers to the process of learning a parametric mapping from the raw input data domain to a feature vector or tensor, in the hope of capturing and extracting more abstract and useful concepts that can improve performance on a range of downstream tasks. Often the input domain has a high dimensional space (images, video, sound, text) and the encoded representations reside in a manifold of a much lower dimensionality. While all *dimensionality reduction* methods convert high-dimensional inputs to a lower-dimensional representation, some of these methods do not learn a mapping that meaningfully generalises on new data samples, and that is what representation learning does.

In the early day of Machine Learning (ML), much effort was spent on designing data transformation and preprocessing pipelines, and learning was only used to make a shallow decision based on extracted features. One of the key ingredients in the success of deep learning is the ability to learn and extract through deep layers some useful features from data. The increase in available computation and labelled datasets has enabled the paradigm shift from using hand-designed feature extractors to learned feature extractors. As a result, the focus in research also shifted from feature-engineering to architecture-engineering. Research into deep learning architectures has exploded in recent years and has matured into a few core principles and

The associate editor coordinating the review of this manuscript and approving it for publication was Shagufta Henna.

  

building blocks e.g convolution layer for local data, recurrent layer for sequential data, and attention layer for set data.

As a goal, the task of explicitly learning a good representation in comparison to implicitly learning a good representation to optimise performance for a task, can be tricky. Firstly, it is not entirely clear what makes a good representation. Based on the analysis by Bengio, Courville, and Vincent [9], a good representation has the properties of local *smoothness* of input and representation, is *temporally and spatially coherent* in a sequence of observations, has *multiple, hierarchically-organised* explanatory factors which are *shared* across tasks, has *simple dependencies* among factors and is *sparsely* activated for a specific input.

From these criteria the field of Representation Learning, especially in the Deep Learning circle, has itself developed a number of core principals used to learn a good representation and these are:

- **Distributed**: Representations that are *expressive* and can represent an exponential amount of configuration for their size. This is in contrast with other types of representations such as one-hot encoding, learned by many clustering algorithms;
- **Abstraction and Invariant**: Good representations can capture more *abstract* concepts that are *invariant* to small and local changes in input data;
- **Disentangled representation**: While a good representation should capture as many factors and discard as little data as possible, each factor should be as *disentangled* as possible. Aside from promoting feature re-use in learning systems, it can also be beneficial for other purposes such as explainability.

While distributed representation and abstraction can be achieved to some degree through deep network architectures, invariant and disentangled representations are harder to achieve and are usually implicitly learned with the task. A family of methods collectively called *Contrastive Learning* offers a simple method to encode these properties within a learned representation.

In this paper, we formulate and discuss a *Contrastive Representation Learning* (CRL) framework, which potentially represents another paradigm shift from *architecture-engineering* to *data-engineering*.

Even though contrastive learning has become prominent in recent years due to the success of large pre-trained models in the fields of *natural language processing* (NLP) and *computer vision* (CV), the seminal idea dates back at least to the 1990s [8], [11]. Furthermore, the development into its current form has spanned over multiple sub-fields and application domains, which can make understanding it challenging although the core intuition behind its operation has remained unchanged. In addition, due to the recent successes of contrastive learning in instance discrimination self-supervised learning, it is often incorrectly regarded merely as another self-supervised learning technique, which does not do justice to the generality of contrastive methods.

With the recent surge in interest in Contrastive Learning methods, there is much published work associated with contrastive learning but without a proper framework to analyse this work, it can be hard to understand the novelties and trade-offs of new methods. This paper proposes a simple yet powerful framework that can be used to categorise and explain in simple terms, the progress in this sub-field, ranging from supervised to self-supervised methods, in multiple application and input domains including images, videos, text and audio and their combinations. To the best of our knowledge, this is the first paper to survey the specific history and recent development of the contrastive approach in a wide range of domains.

In summary, these are the contributions of this paper:
- We propose a simple framework to understand and explain the workings of contrastive representation learning;
- We provide a comprehensive survey of its history and development, and a taxonomy for each of the framework components as well as a summary of conceptual advancements spanning over many sub-fields;
- We study and make connections between the contrastive approach and various other methods;
- We present the application of contrastive learning in various application domains and tasks;
- We analyse the current limits and discuss future research directions.

The rest of the paper is organised as follows. In the next section, we present an introduction and an overview of what contrastive learning is with an emphasis on contrastive learning of representations. That is followed by a taxonomy of how we see contrastive learning starting with a formal framework description and then presenting various ways in which the field can be divided based on similarity, encoders, transform heads and loss functions. Section IV then presents a variety of data domains and problem topics to which contrastive learning has been applied, covering applications in language, vision, audio, graph-structured data, multi-modal data and other areas. We then present a discussion of several topical issues with an emphasis on future outlook, and a concluding section completes the paper.

## II. WHAT IS CONTRASTIVE LEARNING ?

We now present an overview of different representation learning approaches and an intuitive introduction to contrastive learning with a concrete example of the *Instance Discrimination* task in learning self-supervised visual representations.

### A. REPRESENTATION LEARNING
#### 1) GENERATIVE AND DISCRIMINATIVE MODELS
In the machine learning literature, approaches to learning representations of data are often divided into two main categories: generative or discriminative modelling. The process of extracting representations, or inferring latent variables from a probabilistic view of a dataset, is often called **inference**. While both approaches assume that a good representation will capture the underlying factors that explain variations in the data $\mathbf{x}$, they differ in the process of learning these representations. Generative approaches learn representations by modelling the data distribution $p(\mathbf{x})$, for example: all the pixels in an image. It is based on the assumption that a good model $p(\mathbf{x})$ that can generate realistic data samples, must also in turn capture the underlying structure related to the explanatory variables $y$. Evaluating the conditional distribution $p(y|\mathbf{x})$ for some discriminative tasks on variable $y$ can then be obtained by using Bayes' rule.

Discriminative approaches to learning representations on the other hand learn a representation by directly modelling the conditional distribution $p(y|\mathbf{x})$ with a parametrised model that takes as input the data sample $\mathbf{x}$ and outputs the label variable $y$. Discriminative modeling consists of an inference step that infers the values of the latent variables $p(\mathbf{v}|\mathbf{x})$, and then directly makes downstream decisions from those inferred variables $p(y|\mathbf{v})$.

Discriminative models have some advantages when compared to generative models. Modelling the distribution for the set of data $\mathbf{x} \in \mathcal{X}$ is computationally expensive and is not necessary in order to extract representations. If the goal is only to learn a mapping to a lower dimension representation, the generation process in a generative model can be considered wasteful. In addition, the task of learning a good decoder/generator can be entangled with the task of learning a good feature encoder. The objective functions of generative models are also more expensive to evaluate and harder to design since they usually operate in the input space.

### 2) SUPERVISED AND UNSUPERVISED LEARNING

Until recently, the most successful applications of deep learning belonged to the class of supervised learning methods, where a representation is directly learned by mapping from the input to a human-generated label i.e. in training data pairs $(x, y)$, to optimise an objective function. Earlier paradigms involving pre-training layer-wise unsupervised models provided little or no benefit in an end-to-end supervised setting. As the performance of deep learning can scale upwards with the amount of data and the model size [55], the need for labelled data has been identified as an impeding factor in scaling deep networks. Labelling data has its own set of disadvantages such as being time-consuming and expensive, as well as carrying privacy concerns when labelling is out-sourced to a third-party, for example in medical data, as well as potentially injecting annotators' biases through the labelling process.

Previously, most unsupervised representation learning methods belong to the class of generative models [32], [58]. While generative models provide a general objective to learn a low-dimensional representation, they are computationally expensive and are also limited by the ability to model the dependencies between input dimensions.

Until recently, most discriminative approaches to learning representations are a type of supervised learning. Some newer works under the term "self-supervised" learning aim to learn useful representations without labels using discriminative modelling approaches. These methods have shown great success when used for transfer learning, surpassing supervised pre-trained models in multiple downstream tasks, in both computer vision and natural language processing applications. Since a self-supervised discriminative model does not have labels corresponding to the inputs like its supervised counterparts, the success of self-supervised methods comes from the elegant design of the pretext tasks to generate a pseudo-label $\hat{y}$ from part of the input data itself [26], [118].

Among the most successful of the recent self-supervised approaches to learning visual representations, a subset of these termed "contrastive" learning methods have achieved the most success.

### 3) OBJECTIVE AND EVALUATION OF REPRESENTATIONS

As distinct from supervised tasks where we can optimise for the goal directly, there is no straightforward objective for learning good representation that we can measure. A good representation can be useful to improve the performance of downstream tasks, but this objective requires prior knowledge of the task we ultimately want to optimise for. In other cases, where a lot of tasks share the same input data (e.g image classification / detection / segmentation), we can argue that a good representation captures the inherent properties of the underlying data and should be shared across those tasks. However the representations for these tasks can sometimes be in conflict, especially when the capacity of the model is not enough, and there is a trade-off between performance in one task and generalisability across multiple tasks.

In some other cases where a representation is learned through a proxy task, the performance of the proxy task can also be a proxy performance measure for the representation, such as a generative model measuring the reconstruction error, fidelity and diversity of generated samples. This is the case for contrastive representation learning, where the quality of the representation is approximated by how well the representation separates similar and dissimilar samples.

Moreover, sometimes learning a good representation can be an effective way to study the inherent characteristics of the data itself, without the need to perform any particular task.

### B. CONTRASTIVE REPRESENTATION LEARNING

Intuitively, contrastive representation learning can be considered as learning by comparing. Unlike a discriminative model that learns a mapping to some (pseudo-)labels and a generative model that reconstructs input samples, in contrastive learning a representation is learned by comparing among the input samples. Instead of learning a signal from individual data samples one at a time, contrastive learning *learns by comparing* among different samples. The comparison can be performed between positive pairs of "similar" inputs and negative pairs of "dissimilar" inputs.

Unlike supervised methods where a human annotation $y$ is needed for every input sample $\mathbf{x}$, contrastive learning approaches only need to define the similarity distribution in order to sample a positive input $\mathbf{x}^+ \sim p^+(\cdot|\mathbf{x})$, and a data distribution for a negative input $\mathbf{x}^- \sim p^-(\cdot|\mathbf{x})$, with respect to an input sample $\mathbf{x}$. The goal of contrastive learning is very simple: the representation of "similar" samples should be mapped close together, while that of "dissimilar" samples should be further away in the embedding space. Thus by contrasting between samples of positive and samples of negative pairs, representations of positive pairs will be pulled together while representations of negative pairs are pushed far apart.

In the self-supervised setting, instead of deriving a pseudo-label from the pretext task, contrastive learning methods learn a discriminative model on multiple input pairs, according to some notion of similarity. Similar to other self-supervised pretext tasks, this definition of similarity can be defined from the data itself, and thus can overcome a limitation encountered in supervised learning settings where only a finite number of label pairs are available from the data. While some self-supervised methods need to modify the model architecture during learning (such as in [118]), contrastive methods are much simpler where no modification to the model architecture is needed between training and fine-tuning to other tasks.

If additional labels are provided, these can also be integrated into the definition of similarity and dissimilarity of the contrastive framework as well. By defining the similarity and dissimilarity distribution on the dataset level instead of individual data samples, contrastive methods alleviate the need for a labelled dataset while providing a mechanism to specify the desired invariant / covariant properties of the learned mapping. Thus contrastive learning methods provide a simple yet powerful approach to learning representations in a discriminative manner in both supervised or self-supervised setups.

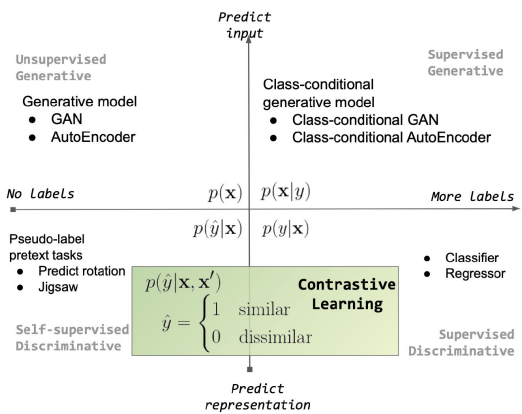

**FIGURE 1.** Contrastive learning in the Generative-Discriminative and Supervised-Unsupervised spectrum. Contrastive methods belong to the group of discriminative models that predict a pseudo-label of *similarity* or *dissimilarity* given a pair of inputs.

Figure 1 illustrates the family of contrastive methods along generative-discriminative and supervised-unsupervised axes.

### C. EXAMPLE: INSTANCE DISCRIMINATION

Along the lines of an exemplar-based classification task [26], which treats each image as its own class, Instance Discrimination [110] is a popular self-supervised method to learn a visual representation and has succeeded in learning useful representations that achieve state-of-the-art results in transfer learning for some downstream computer vision tasks [43], [69]. Based on the simple formulation proposed in SimCLR [16], in this section we will describe the Instance Discrimination task as a simple form of contrastive learning, as illustrated in Figure 2.

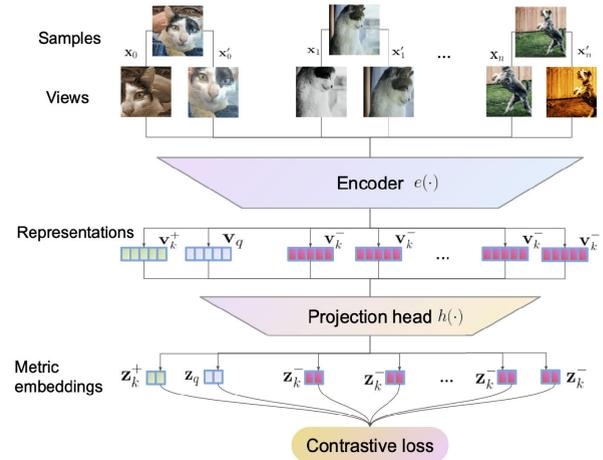

**FIGURE 2.** Contrastive learning in the Instance Discrimination pretext task for self-supervised visual representation learning. A positive pair is created from two randomly augmented views of the same image, while negative pairs are created from views of two different images. All views are encoded by the a shared encoder and projection heads before the representations are evaluated by the contrastive loss function.

The image-based instance discrimination pretext task learns a representation by maximising agreement of the encoded features (embeddings) between two differently augmented views of the same images, while simultaneously minimising the agreement between views generated from different images. To avoid the model maximising agreement through low-level visual cues, views from the same images are generated through a series of strong image augmentation methods.

- Let $\mathcal{T}$ be the set of *image transformation* operations where $t, t' \sim \mathcal{T}$ are two different transformation operators independently sampled from $\mathcal{T}$. These transformations could be random *cropping* and *resizing*, *blur*, *color distortion* or *perspective distortion*, etc. A $(\mathbf{x}_q, \mathbf{x}_k)$ pair of query and key views is positive when these two views are created by applying different transformations on the same image $\mathbf{x}$: $\mathbf{x}_q = t(\mathbf{x})$ and $\mathbf{x}_k = t'(\mathbf{x})$, and is negative otherwise.
- A *feature encoder* $e(\cdot)$ then extracts the feature vectors from all the augmented data samples $\mathbf{v} = e(\mathbf{x})$. There is no restriction on the choice of the encoder but a ResNet [42] model is usually used for image data because of its simplicity. The representation $\mathbf{v} \in \mathbb{R}^d$ in this case is the output of the average pooling layer of Resnet.
- Each representation $v$ is then fed into a *projection head* $h(\cdot)$ comprised of a small multi-layer perceptron (MLP) to obtain a metric embedding $\mathbf{z} = h(\mathbf{v})$, where $\mathbf{z} \in \mathbb{R}^{d'}$ with $d' < d$ is in a lower dimensional space than the representation **v**. This projection head can be as simple as a one-layer MLP using a non-linear activation function. All the vectors are then normalised to be unit vectors.

- A batch of these metric embedding pairs $\{(\mathbf{z}_i, \mathbf{z}'_i)\}$, with $(\mathbf{z}_i, \mathbf{z}'_i)$ represents the metric embeddings from two augmented versions $(\mathbf{x}_q, \mathbf{x}_k)$ of the same image, are then fed into the *contrastive loss* function which encourages the distance in the metric embedding of the same pair to be small, and the distances of embeddings from different pairs to be large. The non-parametric classification loss [110] and its variants, such as InfoNCE [77] and NT-Xent [16] is a popular choice for the contrastive loss function, which for the $i$-th pair has the general form:

$$\mathcal{L}_i = -\log \frac{\exp(\mathbf{z}_i^\top \mathbf{z}'_i / \tau)}{\sum_{j=0}^{K} \exp(\mathbf{z}_i \cdot \mathbf{z}'_j) / \tau)} \quad (1)$$

where $\mathbf{z}^\top \mathbf{z}'$ is the dot product between two vectors and $\tau$ is a temperature hyper-parameter that controls the sensitivity of the product. The sum in the denominator is computed over one positive and $K$ negative pairs in the same minibatch. Intuitively, this can be understood as a non-parametric version of $(K + 1)$-way softmax classification [110] of $z_i$ to the corresponding $z'_i$.

In order to minimise the InfoNCE loss function in Eq. (1), the dot product in the numerator measuring the similarity of representation from the same pair is maximised, while the similarity of all negative pairs in the denominator, is minimised.

When the contrastive training phase is done, the projection head is discarded and the encoder is used as the feature extractor for transfer learning. By combining the predictor or classifier with the representation output of the encoder, they can be fine-tuned on a new task on a target dataset.

Contrastive methods in the instance discrimination task set out to learn a representation that can separate between different instances, while ignoring the meaningless variances introduced by image data augmentation. Because contrastive learning directly maximises similarity between representations of positive similar pairs and minimises that of negative pairs, how those pairs are generated directly determines the invariant properties in the learned representation. The most important components for the success of contrastive pre-training on ImageNet [23] is data augmentation methods. As analysed in SimCLR [16], many contrastive methods perform very poorly without proper augmentations (i.e random crop and color distortion) even for the same set of architectures and losses.

The dataset, data transformations and instance-wise similarity definition combined together in the contrastive learning framework provide a scalable and accessible approach to specifying invariant and covariant properties in the learned representation.

## III. A TAXONOMY FOR CONTRASTIVE LEARNING

Before we present our taxonomy for contrastive learning methods, we first formally describe the contrastive representation learning (CRL) framework in Section III-A. In particular, the CRL is a general framework that can be used to succinctly describe a variety of contrastive learning methods ranging from self-supervised to supervised and covering images, videos, audio, text and more. We use this framework to introduce a comprehensive taxonomy for the components of contrastive methods in Sections III-B, III-C, III-D and III-E.

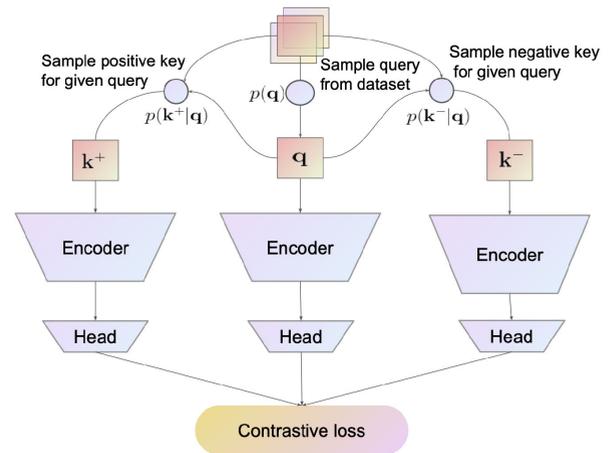

**FIGURE 3.** Overview of the Contrastive Representation Learning framework. Its components are: a similarity and dissimilarity distribution to sample positive and negative keys for a query, one or more encoders and transform heads for each data modality and a contrastive loss function evaluate a batch of positive and negative pairs.

### A. THE CONTRASTIVE REPRESENTATION LEARNING FRAMEWORK

The general CRL framework, illustrated in Fig. 3 builds on top of the work of Chen *et al.* [16], which describes a simple contrastive self-supervised framework to learn visual representations in the context of an Instance Discrimination task (see Section II-C). As distinct from [16], we generalise this framework beyond the image Instance Discrimination task to cover learning representations in a variety of data domains (images, video, audio and text), learning setups (supervised, self-supervised or knowledge distillation) and ways to define the concept of similarity. Specific choices of the similarity distribution, encoders and heads as well as contrastive loss functions allows the CRL framework to encompass arbitrary contrastive learning methods. More importantly, it enables a clear understanding of most of the contemporary work and sheds light on the limitations and the promising directions ahead.

In the following and throughout the rest of the paper, we adopt the metaphor of *query* and *key* similar to [43], by considering the problem of similarity matching as a form of dictionary look-up.

We will use the symbols *q* and *k* to represent the *query* and *key* for either the input sample **x**, the representation **v** or the metric embedding **z** depending on context. When we need to be specific, the corresponding symbols **x**, **v**, **z** with superscript $\cdot^q$, $\cdot^k$ for query and key will be used.

*Definition 1 (Query, Key):* *Query* and *key* refer to a particular view of an input sample $\mathbf{x} \in \mathcal{X}$. Together they form a positive or negative pair $(\mathbf{q}, \mathbf{k})$ depending on whether the query and key are considered similar or not.

In the Instance Discrimination task, query and key views are a randomly transformed version of an image $t(\mathbf{x})$ in the data set $\mathcal{X}$.

*Definition 2 (Similarity Distribution):* A *similarity distribution* $p^+(\mathbf{q}, \mathbf{k}^+)$ is a joint distribution over a pair of input samples that formalises the notion of similarity (and dissimilarity) in the contrastive learning task. Distinct from other machine learning methods where the data distribution is defined over a single input sample $p(\mathbf{x})$, the *similarity* required by contrastive methods takes input from the joint distributions of pairs of samples $p(\mathbf{q}, \mathbf{k})$.

A key is considered positive $\mathbf{k}^+$ for a query $\mathbf{q}$ if it is sampled from this similarity distribution and is considered negative $\mathbf{k}^-$ if it is sampled from the dissimilarity distribution $p^-(\mathbf{q}, \mathbf{k}^-)$. In some tasks, the dissimilar data distribution may not be explicitly defined but implicitly given as the distribution of any pair that is not sampled from the similarity distribution.

Similar to other representation learning problems, the focus of contrastive learning is in learning from a high-dimensional input space $\mathcal{X}$, which depends on the domain and can be a tensor representing audios, images, videos or texts.

Combining the data distribution $p(\mathbf{x})$, the definition of similarity $p^+(\mathbf{q}, \mathbf{k})$ and dissimilarity $p^-(\mathbf{q}, \mathbf{k}^-)$, different properties of the learned representation can be specified, as illustrated in Figure 4.

In practice, queries and keys are not necessarily sampled jointly but the query can be sampled first from the data distribution $\mathbf{q} \sim p(\mathbf{x})$ where then the corresponding positive and negative keys are sampled from the conditional distributions $\mathbf{k}^+ \sim p^+(\cdot|\mathbf{q})$ and $\mathbf{k}^- \sim p^-(\cdot|\mathbf{q})$.

In the Instance Discrimination task, the similarity distribution is defined over any pair that are transformed from the same input samples $\mathbf{q}, \mathbf{k} \sim p^+(\cdot, \cdot)$ if $\mathbf{q} = t(\mathbf{x})$ and $\mathbf{k} = t'(\mathbf{x})$ for 2 different random transformations $t$ and $t' \in \mathcal{T}$.

*Definition 3 (Model):* We refer to the combination of all modules with parameters in a contrastive learning method as the **model** $f(x; \theta) : \mathcal{X} \to \mathbb{R}^{|\mathcal{Z}|}$ and its parameters collectively as $\theta$.

The model can be decomposed further into a base encoder and a transform head.

*Definition 4 (Encoder):* The features *encoder* $e(\mathbf{x}; \theta_e) : \mathcal{X} \to \mathcal{V}$ with parameters $\theta_e$ learns a mapping from the input views $x \in \mathcal{X}$ to a representation vector $v \in \mathbb{R}^d$. This network (when trained via contrastive learning) can be used to generate features (or inputs) to leverage the learned representations in other tasks (e.g. as input when learning another model for an image classification task), or to have layers stacked on top (e.g. fully connected, softmax) where the network can be fine-tuned to the new task.

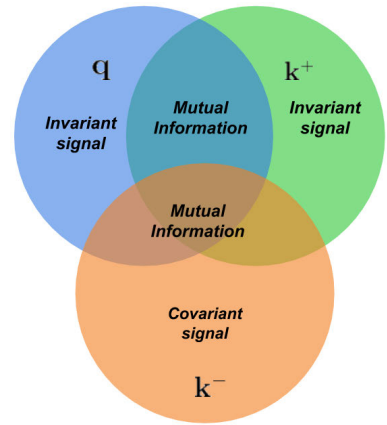

**FIGURE 4.** An intuitive diagram represents the learning signal captured by the contrastive loss through the query, positive and negative keys. Contrastive methods allow the desired invariances to be specified through the similarity and dissimilarity distributions. Each circle represents the information signal contained in each view. The signal that is not mutual between query and positive keys are invariant features, since their representations are made as similar as possible. The signal that is not mutual between the negative key and the query or positive keys are covariant features, since these representations must be able to distinguish between those to minimise similarity to the negative key.

*Definition 5 (Transform Head):* Transform heads $h(\mathbf{v}; \theta_h) : \mathcal{V} \to \mathcal{Z}$ parameterised by $\theta_h$, are modules that transform the feature embedding $\mathbf{v} \in \mathcal{V}$ into a metric embedding $\mathbf{z} \in \mathbb{R}^{d'}$.

Depending on the specific application, the transform heads can be used to aggregate information from multiple representation vectors or used to project it down to a lower-dimensional space before the contrastive loss.

*Definition 6 (Contrastive Loss):* A contrastive loss function operates on a set of metric embedding pairs $\{(z, z^+), (z, z^-)\}$ of the query, positive and negative keys. It measures the similarity (or distance) between the embeddings and enforces constraints such that the similarity of positive pairs are high and the similarity of negative pairs are low. To attain small distances between the embeddings of positive pairs in the metric space, representations will become **invariant** to irrelevant differences in the input space of positive pairs, while simultaneously learning the **covariant** representation between negative pairs to explain for the large distance in the metric space.

### B. A TAXONOMY OF SIMILARITY

Contrastive Learning revolves around learning a mapping from different views of the same *scene*, or *context* into the same region of a representation space, which is formalised through the similarity distribution. The key to an effective contrastive learning task is to design the similarity distribution such that positive pairs are very different in the input space yet semantically related, and a dissimilarity distribution such that negative pairs are similar in the input space but semantically different. Despite the recent popularity of self-supervised contrastive learning, contrastive learning in general is agnostic to the supervised / unsupervised paradigm. Depending on whether any human labels $y$ are used in defining those joint distribution (e.g. $k \sim p(\cdot|q, y)$), the method then becomes a supervised or self-supervised contrastive learning task.

Depending on the end goals there can be many notions of similarity and dissimilarity, which is a strong point of contrastive methods, but it also makes it difficult to provide a taxonomy that captures all these variations. However, there are some general principles that are usually the underlying assumptions behind how similarity and dissimilarity is constructed, which we now examine.

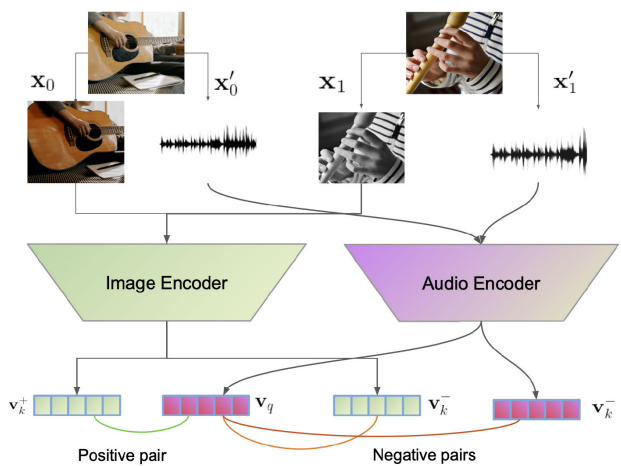

**FIGURE 5.** Illustration of learning similarity between multiple modalities. Each modality has an encoder and the representations extracted by different encoders are contrasted with each other to learn a joint embedding space.

#### 1) MULTISENSORY SIGNALS
One direct approach to have multiple views of the same context is to record the information with multiple sensors. These sensors can be of the same modality (e.g. two cameras recording the same scene from different angles), or of different modalities (e.g. audio and image from a video), as illustrated in Fig. 5. Using the natural correspondence between different sensors, the model can learn to be invariant to the low-level details in each sensor input and focus on representing the shared context between them.

Contrastive methods have been used to learn cross-modal representations of visual and textual data in [50], [96]. In the Time-Contrastive Network [91], a visual representation is learned by pulling the representation of two simultaneous views from *different cameras* of the same scene, while pushing apart *frames taken from far away in time but from the same video*. This leads to a representation space that is invariant to viewpoints while being sensitive to changes in time.

#### 2) DATA TRANSFORMATION
If synchronous data from multiple sensors is not available (e.g. a single-modality dataset like ImageNet), the most simple yet effective approach to generating different views of the same scene is to use a hand-crafted transformation function operating on the input data domain. Designing and implementing such semantic-preserving transformations requires prior knowledge, but this knowledge is defined once for the entire dataset or data collection pipeline, and can be dynamically applied to individual samples at run time.

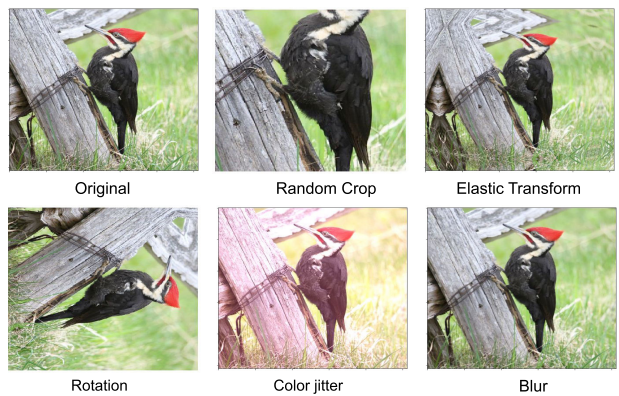

**FIGURE 6.** Illustration of some common image augmentation methods. Different views from a random set of augmentations of the same images are usually considered positive pairs.

For visual data, image augmentation methods such as *lighting or color distortion*, *cropping and padding*, *adding noise and blur*, *rotation and perspective transform* etc. are efficient methods to transform pixels while preserving the semantic meaning of an image's content such as its class labels. An example of these data transformations techniques on image can be seen in Fig. 6. Destroying low-level visual cues by image augmentation forces the contrastive method to learn a representation invariant to those changes in the inputs. These techniques have been widely used in supervised learning to learn invariant features and to increase the robustness of the resulting models. The recent wave of instance discrimination contrastive methods have demonstrated that the same representation can be learned from these augmentation techniques without the need for a class label [16], [43], [69], [110], [117].

For natural language text data, Fang *et al.* [29] transform a sentence using a back-translation method to create a slightly different sentence that has the same semantic meaning as the original one to form a positive pair. Back-translation uses two machine translation models to translate a sentence into a target language and back to the source language. The randomness from the two translation models will yield a sentence in the source language that is slightly different from the original sentence.

For program code data, ContraCode [51] uses various source-to-source transformation methods from the compiler literature such as *variable renaming, identifier mangling, reformatting, beautification, compression, dead-code* *insertion / elimination*, etc. to construct semantically similar code snippets that share the same functionality. Learning to map these textually different but functionally equivalent programs to the same feature vector allows the model to learn a function representation space that is predictive of equivalent programs.

For audio data, some augmentation methods such as *warping*, *frequency and temporal masking* in the Mel spectrogram format could be used to create different version of the same audio data, as in [73].

### 3) CONTEXT-INSTANCE RELATIONSHIP

Another approach to extracting similar views of the same scene is by exploiting the context-instance relationship from a sample representation. Generally, we want to learn a representation that captures the entire context, i.e the global information about a scene. That context can usually be decomposed further into parts, each containing a subset of the scene's information that is local to each subset.

Explicitly constraining the representation of the parts (local features) to be similar to the representation of the whole (global features), while being different from the representation of other views is a clever approach to defining similarity. Contrasting between the representation of local features versus global features can encourage the model to learn important features that present in the local views, while ignoring noise features which occur only in those local inputs. Representation from local features is thus encouraged to capture meaningful information relevant to the whole context, while global features are encouraged to capture as much detail from the local instances as possible.

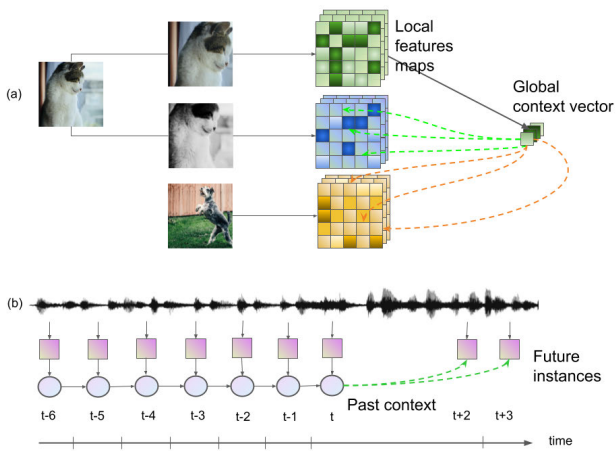

**FIGURE 7.** Illustration of extracting query and keys using the context-instance relationship. In *a)*, the context is a global summary vector of the entire image, while the instances are the local features in the set of intermediate feature maps. In *b)*, the past context is aggregated with a RNN contextualisation head and the instance are representations of future time steps.

Figure 7a describes the approach taken in Deep InfoMax (DIM) [46], where an image is encoded into a global feature vector and also into a feature map corresponding to spatial patches of pixels in the original image. The global feature and local features in the feature map of the same images then form positive pairs, while global features with local features from other images are considered negative pairs.

Global features can also be constructed from videos in the temporal dimension, as in Fig. 7b. In Contrastive Predictive Coding (CPC) [77], context features are constructed as a summary of past input segments, and then contrasted with local features from a future time step. Contrastive learning to predict the correct future from the past context in this way can be thought of as an instantiation of the predictive coding theory.

### 4) SEQUENTIAL COHERENCE AND CONSISTENCY

In addition to the context-instance feature relationship, exploiting the spatial or temporal coherence and consistency in a sequence of observations is another approach to defining similarity in contrastive learning. This method works for a data domain that can be decomposed into a sequence of smaller units, such as an image into a sequence of pixels, or a video into a sequence of frames, etc. The representation of continuous views in a sequence is considered as a positive pair while discontinuous and far away pairs in the same sequence or different sequences are considered negative pairs. This approach uses the *slowness assumptions* in representation learning, which states that important features are the ones that change slowly over a sequence of observations. Therefore, by learning invariant, slowly changing features in a sequence, a model will learn to extract the most important features in the data, illustrated in Fig. 8.

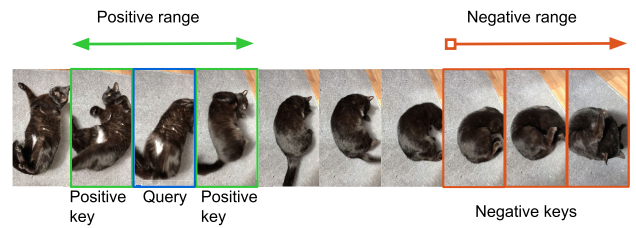

**FIGURE 8.** Illustration of sampling query and keys using the sequential coherence property of video data. The positive keys are defined as frames inside a small window surrounding the query frame. The negative keys are frames from the same video but are far away in time to the query.

Rather than using simultaneous videos with multiple viewpoints as in Time-Contrastive Network (TCN) [27], [91] uses a *multi-frame TCN* that exploits the temporal coherence property of video and applies contrastive learning on a sequence of frames, where frames inside a time-window are positive to each other, and pairs from with a frame outside the window are considered negative.

In addition to the hand-crafted transformations described in Section III-B2, the temporal coherence of video frames can also provide a natural source of data transformations. In a video, an object can undergo a series of transformations such as object deformation, occlusion, changes in viewpoint and lighting. These methods have been used in [82], [106] to learn representations of objects from videos without any additional labels.

#### 5) NATURAL CLUSTERING

Clustering is the process of finding high-level semantics for groups of instances features according to some distance measure in the embedding space. Natural clustering refers to the assumption that different objects are naturally associated with different categorical variables, where each category occupies a separate manifold in a representation space. The distance between different clusters loosely represents the similarity between categories. This assumption is consistent with how humans naturally categorise and name different groups of objects, and is an important assumption in unsupervised learning, manifesting itself in various clustering algorithms such as K-Nearest neighbors. Semantic class labels in classification problems are also an instance of this assumption where the number of clusters and the names for these clusters are given by human annotators. Each cluster represents a high-level semantic concept and together the set of clusters provide overall structure to the data manifold.

Contrastive learning induces a metric in the embedding space where positive pairs have smaller distances between them and negative pairs have large distance, based on a semantic definition of similarity. In contrast to clustering methods which enforce the cluster assumption in a top-down fashion, contrastive methods enforce local smoothness between positive pairs thus organising the embedding manifold from the bottom up. Since contrastive learning and clustering methods essentially encode the same assumption but from different directions, the combination of contrastive methods from bottom-up and clustering approaches from top-down are a promising approach which complement each other's advantages. Figure 9 demonstrates this idea of combining contrastive learning with clustering methods.

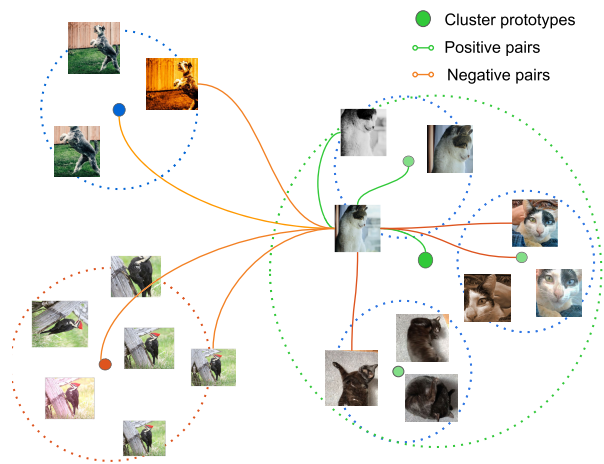

**FIGURE 9.** Illustration of contrastive methods on clusters. In addition to individual sample's vector, there can also have cluster prototypes with different levels of granularity. Contrastive loss can operate on both the sample and cluster level.

Many different methods have tried to use contrastive methods to learn invariant properties while supplementing higher-level semantic information to the contrastive framework using clustering methods, such as *Prototypical Contrastive Learning* (PCL) [63], or *Swapping Assignment between multiple views* (SwAV) [14]. In [56], the class labels for a supervised learning task are provided as cluster information to improve on the traditional self-supervised instance discrimination task.

### C. A TAXONOMY OF ENCODERS

In contrastive representation learning, a learned mapping from inputs to the embedding space needs to satisfy two purposes: mapping to a general and powerful representation of the input data, and an efficient and effective embedding that allows measurement of the distances between samples. We divide the model in our contrastive representation learning framework into two components based on recognising the purpose and functionality of each component i.e. the base encoder and transformation head. The purpose of the encoder is to learn a good mapping from inputs to a general representation space, while the transform heads, depending on the specific choice of similarity, will transform one or multiple representations to a metric embedding for computing a similarity metric. In practice there may be no distinction between the base encoder and the head from a technical point of view as they are just layers of a deep network, stacked on top of each other and jointly optimised through back-propagation with gradient descent but they are functionally distinct, hence the separation.

In this sub-section we focus on a taxonomy of the base encoders. While contrastive learning is general and not restricted to any particular form of encoder, some specific types of encoder and the interactions among them will enable different behaviours for the downstream transform heads and contrastive loss. For each data modality, an appropriate encoder architecture is chosen, so the taxonomy for the encoder will be based on how they are updated with respect to the gradient from the contrastive loss during training.

#### 1) END-TO-END ENCODERS

End-to-end encoders represent the most simple method both conceptually and technically, where the encoders for the queries and keys are updated directly using gradients back-propagated with respect to the contrastive loss function. Since all encoders are updated end-to-end, this can impose a significant requirement on memory. Therefore if the query and keys are of the same data modality, their respective encoders are usually shared with each other so only one copy of the encoder needs to be stored in memory. This way, both the representation for the queries and keys can be efficiently batch-computed in one single forward pass. However, encoding both the queries and keys end-to-end still requires storing the hidden activations and representation on a Graphical Processing Unit's Video Memory (GPU's VRAM), which will limit the batch size for calculating the contrastive loss.

### 2) ONLINE-OFFLINE ENCODERS

The online-offline encoders approach alleviates the memory requirement of end-to-end encoders for storing all the queries and keys in a GPU's memory by using an additional offline encoder, which is not updated online by gradient descend directly but updated offline from the online network. In this way, the feature vectors and the hidden activations computed by the offline encoder are not stored on the VRAM. Therefore with this approach, contrastive methods can scale up the number of positive and negative pair comparisons in a batch, independent of the GPU's memory limit.

There are generally two ways to update the offline network, either by using a *past checkpoint* or via a *momentum-based weighted average* mechanism from the online encoder.

Wu *et al.* [110] decoupled the batch size from the number of negative pairs by storing a detached copy of representations of the entire dataset into a separate *memory bank*. The representations stored in this memory bank are later randomly sampled to serve as the keys, while the queries are encoded by the online network from two different transformations of the same images. The representations computed from the online encoder for the queries are then stored in the memory bank to be used as the keys for the next epoch. This approach effectively uses an online encoder's checkpoint from the previous epoch as the offline encoder for negative keys in the current epoch, with a memory mechanism to avoid redundant computation.

*Momentum Contrast* (MoCo) [43] further reduces the need to store an offline representation of the entire dataset in the memory bank through the use of a dynamic *memory queue*. The offline momentum encoder is a copy of the online encoder, with parameters being an exponentially-weighted average of that of the online encoder. At every iteration, the latest batch of feature vectors from the momentum encoder are pushed to the memory queue while the oldest batch of features are discarded from the queue. The momentum queue therefore retains a more consistent set of negative keys to the queries and keys encoded online, compared to the memory bank's feature vectors which are only updated once per epoch.

### 3) PRE-TRAINED ENCODERS

Another case of not having to keep an encoder in the GPU's memory is when an encoder is already pre-trained and does not need to be updated at all. This usually happens in cross-modal learning or in a knowledge distillation setting, where contrastive methods are used to learn a mapping to the same representation space of another encoder. This approach decouples the task of learning representation for each modality and can simplify the learning task of each encoder while still leveraging the information shared from different data modalities.

In [96], Sun *et. al.* used a pre-trained *Bidirectional Encoder Representations from Transformers* (BERT) [24] to process discrete automatic speech recognition tokens, while training a separate video BERT model to process continuous video features.

In a knowledge distillation setting, a large pre-trained "teacher" network with frozen weights is used to encode the keys, while a smaller "student" network tries to match the query representation to positive keys from the teacher network. This is a special case where even though the query and key are of the same modality, they are encoded using different encoders. Contrastive Representation Distillation (CDR) [100] uses a large, pre-trained teacher network as the encoder for both the positive and negative keys, while the queries are encoded by a small network learned to match the representation of the teacher network.

### D. A TAXONOMY OF TRANSFORM HEADS

The distinction between the base encoder and the transform heads is to separate the ultimate goals of learning a good representation from that of learning an embedding that is efficient and effective for computing and maximising the similarity metric. Entangling the main task of learning a representation and the pretext task of learning a similarity metric can leads to unwanted results, such as by only focusing on maximising the similarity between positive samples, the representation is forced to discard potentially useful information. The introduction of an explicit transform head above encoders is a recent development in contrastive representation learning. Prior to the introduction of the transform heads, many methods trained a standard encoder and then performed a comparison of which layers are best suited to use as representation for transfer learning to some downstream tasks. The result was that for most tasks, one of the hidden layers gave the best performance when using as a representation for transfer learning or fine-tuning with a downstream classifier.

With the separation from the base encoder and transform heads, it is now also possible to train the same representation from the base encoder with multiple transform heads for different contrastive objectives.

Depending on the specific choice of data similarity (see Section III-B) and its purpose, we categorise transform heads into three types namely *projection*, *contextualisation* and *quantisation* heads which we now describe in turn.

### 1) PROJECTION HEADS

While the representations (the output of encoders) are of a lower dimensionality to the input dimensions, it can still take a relatively large computational effort to measure the similarity distance between representations. The simplest type of transformation serves as a bridge between different vector spaces. These projections can be a simple linear transformation or a non-linear MLP. With the projection head, the dimensionality for the representation $\mathbf{v}$ can be larger than the dimensionality of the metric embedding $\mathbf{z}$, so that more information can be retained in the representation while also allowing for efficient computation of the similarity metric in the space of $\mathcal{Z}$.

The early contrastive methods that report transfer learning results from the best hidden layers are effectively using the base of the network as a feature encoder and the top of the network as the non-linear projection head. In more recent work, [117] explicitly uses a linear and [16] uses a non-linear 2-layer MLP as the projection head after the base encoder.

Instead of projecting the representation of the query and key encoders to a common metric space, a transformation head can also be used to bridge directly from one metric space to another. In [34], in addition to a projection head from representation space to metric space, an additional "prediction" network projects the metric embedding of an online network to the the metric embedding of an offline encoder.

#### 2) CONTEXTUALISATION HEADS

In some settings, the projection heads can be more elaborate than just simply projecting the representation down to a lower dimension. For the task that defines similarity based on the context-instance relationship (Section III-B3), a special kind of transform head is needed to aggregate multiple feature vectors into a contextualised embedding.

In *Contrastive Predictive Coding* (CPC) [77] where similarity is defined from the past-present relationship, a GRU [19] head is applied over previous time steps to aggregate the past information into a contextualised embedding. This is equivalent to an ordered autoregressive head that forces the head to learn generalisable features that are informative when predicting the correct future separate from the incorrect future.

In Deep InfoMax (DIM) [46], where global features are compared with local information in the feature maps, convolution layers with pooling are used to aggregate the feature maps into one single global vector. Similar to DIM, in InfoGraph [97] where contrastive learning is applied on a graph network, a transform function summarises all the patch representations into a single fixed length graph-level representation.

As distinct from the projection head where the representation is only projected down, the contextualised metric embedding $\mathbf{z}$ serves a different function and holds different kinds of information. Depending on the downstream task where the contextual information is helpful or not, the contextualised embedding $\mathbf{z}$ can actually be used instead of, or in conjunction with, the representation embedding $\mathbf{v}$.

#### 3) QUANTISATION HEADS

While a contextualisation head aggregates multiple representations together, a quantisation head is the opposite in that it reduces the complexity of the representation space by mapping multiple representations into the same representation.

For example, wav2vec 2.0 [7] uses a Gumbel-softmax [52] quantisation head to map the continuous audio signal into a discrete set of latent vectors (i.e "code book").

In methods that combine contrastive learning with clustering approaches such as SwAV [14], a Sinkhorn-Knopp algorithm [22] is used as a quantisation head in order to map a representation of individual samples into a soft cluster assignment vector.

### E. A TAXONOMY OF CONTRASTIVE LOSS FUNCTIONS

Contrastive loss is one of the key differences between contrastive methods and other representation learning approaches. The most prominent difference is that in the contrastive loss formulation, the target can be dynamically defined in terms of the metric embedding instead of having fixed targets. While most discriminative models measure loss with respect to a prediction label for example using class labels, and generative models measure loss in the input space (e.g. reconstruction loss), contrastive losses measure the distance, or similarity, between embeddings in the latent space.

All forms of contrastive losses can be generally decomposed into two components: a *scoring function* that measures the compatibility between two vectors and the actual *form* of the loss that enforces minimisation and maximisation given a set of query and key vectors.

Minimising the distance between samples is the ultimate goal of any contrastive loss function. However naively minimising the distances between positive pairs can lead to a catastrophic collapse, e.g. the distances between any pairs can be reduced to zero by making the model $f(\cdot; \theta)$ constant with respect to any input $\mathbf{x}$. To prevent this collapse from happening, the contrastive loss function can explicitly use negative pairs that are forced to have a large distance in the embedding space, or we can implicitly employ other assumptions and architecture constraints. For example, in some recent work such as BYOL [34] or [28], negative pairs are not employed explicitly, and here the authors do not refer to their method as a "contrastive learning" approach. However, we consider all methods that contrast between a query and positive keys to learn similarity as contrastive learning methods, regardless of whether explicit negative pairs or architectural constraints are used to prevent the representation from collapsing.

Given the goal of optimising the distance or similarity score, contrastive loss functions can generally be classified based on their motivation and the specific form of how they are formulated. Below we will discuss the different types of scoring functions and then look at the three major forms of contrastive loss functions.

#### 1) SCORING FUNCTIONS

The scoring function measures compatibility between two vectors either in terms of *similarity* or *distance*. Depending on the specific loss function, for positive pairs either the similarity score is maximised or the distance metric is minimised.

For contrastive losses that operate on the distance notion, usually a simple Manhattan or Euclidean distance (also known as L1 and L2-norm distance) $D(\mathbf{q}, \mathbf{k}) = \|\mathbf{q} - \mathbf{k}\|_2$ is used. Distance-based scoring function are often used in energy-based hinge loss functions (Section III-E2).

On the other hand, scoring functions can measure similarity via a simple dot product $S(\mathbf{q}, \mathbf{k}) = \mathbf{q}^\top \mathbf{k}$ between two vectors. The range of similarity scores in this case is unbounded and dependent on both the orientation and magnitudes of the vectors in the sub-space. Since similarity can be made arbitrarily large by increasing the magnitude, one possible solution is to include a normalisation term for the vector's magnitude $\|\mathbf{z}\|^2$ in the final loss function, as is done in [110]. Another method to get rid of dependency on magnitude is to use the cosine similarity, which is computed as the dot product between two unit vectors $S(\mathbf{q}, \mathbf{k}) = \frac{\mathbf{q}^\top \mathbf{k}}{\|\mathbf{q}\|\|\mathbf{k}\|}$. The cosine similarity is bounded between -1 and 1 for anti-parallel and parallel vectors respectively, and equal to 0 for orthogonal vectors. This is most commonly used as a scoring function in modern contrastive loss functions such as the NT-Xent loss in SimCLR [16]. Another popular option to measure similarity is the bi-linear model $S(\mathbf{q}, \mathbf{k}) = \mathbf{q}^\top \mathbf{A} \mathbf{k}$, in which the matrix $\mathbf{A}$ is learned and can be considered as a linear projection from the sub-space of $\mathbf{q}$ to sub-space of $\mathbf{k}$, before the dot product operation is performed. The original InfoNCE loss [77] uses this bi-linear model as the scoring function.

In the extreme case, the scoring can also be a learned module and be optimised together with the other modules during training, similar to the discriminator network of a GAN [32]. Different from a GAN's discriminator that evaluates one sample at a time, the learned scoring function concatenates multiple metric vectors together as input and measures the correspondence between them. Though it might be thought that a learnable module is better than a hand-crafted scoring function, using a neural network as a scoring function come with disadvantages. The learned discriminator takes up computational resources that are potentially more helpful for the feature encoder. Therefore, a powerful discriminator can make up for poor representation extracted from an encoder by focusing on learning a good discriminator for bad a representation vector instead of learning a useful representation in itself. The learned scoring functions are also often based on the classification objective, whether the two inputs are compatible or not [3]. It does not provide an explicit measurement of distance and similarity in the latent space, which many downstream applications rely on. Therefore in this paper, we mostly focus on methods that uses a contrastive loss with relatively simple scoring functions.

#### 2) ENERGY-BASED MARGIN LOSSES

*Energy-based Models* (EBM) [62] are a general class of models that associate an energy (distance score) with each configuration of the variables to be modelled (pairs of query and keys vectors). Training an EBM involves associating a low energy (small distance) to desired configurations of the variable (positive pairs) and high energy to undesired configurations of variables (negative pairs). Unlike a properly normalised probabilistic model, making the energy for one particular configuration low does not necessarily make energy for other configurations higher. That is why most energy-based models must employ explicit negative comparisons in computing the total loss.

Motivated from EBM, Chopra, Hadsell, and LeCun [20] first introduced and then reformulated in [39] the original "contrastive loss" that uses Euclidean distance $D(\mathbf{q}, \mathbf{k}) = \|\mathbf{q} - \mathbf{k}\|_2$ as the scoring function in the embedding space. To avoid confusion with the general class of all contrastive loss functions, we will refer to this as the "pair loss". The pair loss operates on a pair of query and key, where distance between positive pairs is minimised while the distance between negative pairs should be larger than a given margin, and formally takes the form:

$$\mathcal{L}_{\text{pair}} = \begin{cases} D(\mathbf{q}, \mathbf{k})^2, & \text{if } \mathbf{k} \sim p^+(\cdot|\mathbf{q}) \\ \max(0, m - D(\mathbf{q}, \mathbf{k})^2), & \text{if } k \sim p^-(\cdot|\mathbf{q}) \end{cases} \quad (2)$$

where the margin $m > 0$ acts as a radius around the query, for which only negative keys $\mathbf{k}^-$ within this radius are pushed away from $\mathbf{q}$ and contribute to the total loss value.

While the *pair loss* only requires the distance of negative pairs to be larger than a fixed margin, the *triplet loss* [15], [21], [108] enforces the *relative* distance between positive and negative pairs given in a triplet of *(query, positive key, negative key)*:

$$\mathcal{L}(\mathbf{q}, \mathbf{k}^+, \mathbf{k}^-) = \max(0, D(\mathbf{q}, \mathbf{k}^+)^2 - D(\mathbf{q}, \mathbf{k}^-)^2 + m) \quad (3)$$

While conceptually simple and widely adopted in multiple metric learning applications [49], [90], [105], the pair and triplet losses usually suffer from slow convergence because of the limited interactions between samples. In pair loss, only one comparison to either a positive or negative key is computed for a given query, while triplet loss simultaneously compares the relative distance from a query to one positive and negative key. Mining techniques to find "hard" negative samples to avoid easy pairs that provide no substantial learning signal are essential components of these learning systems. To increase the number of interactions for a query, methods such as *Lifted Embedding loss* [75] and a generalised version of it [45] improved on the margin formulation of triplet loss to take into consideration multiple positive and negative keys for a query within a batch.

#### 3) PROBABILISTIC NCE-BASED LOSSES

A form of contrastive loss can also be motivated from the probabilistic softmax classification problem. Consider the traditional supervised parametric softmax classification objective, the probability that a query is correctly recognised as belonging to the $i$-th class among $n$ classes is

$$p(i|\mathbf{q}) = \frac{\exp(\mathbf{q}^\top \mathbf{w}_i)}{\sum_{j=1}^n \exp(\mathbf{q}^\top \mathbf{w}_j)} \quad (4)$$

where $\mathbf{w}_j$ is a vector specific to the class $i$ in the data set. This vector $\mathbf{w}$ in the parametric formulation of softmax serves as a class prototype and does not allow explicit comparison between representations.

Motivated by this, a non-parametric version for the softmax function that correctly identifies the positive for a given query

P. H. Le-Khac et al.: CRL: A Framework and Review

IEEE Accessfrom a set $\mathcal{K}$ and contains all negative keys with one positive key can be defined as follows:

$$p(\mathbf{k}^+|\mathbf{q}) = \frac{\exp(\mathbf{q}^\top \mathbf{k}^+)}{\sum_{k \in \mathcal{K}} \exp(\mathbf{q}^\top \mathbf{k})} = \frac{\exp(\mathbf{q}^\top \mathbf{k}^+)}{Z(\mathbf{q})} \quad (5)$$

with $Z(\mathbf{q})$ as the normalising constant, or partition function for a given query.

The learning objective is then to maximise the joint probability or equivalently to minimise the negative log-likelihood over the training set:

$$\mathcal{L}(\mathbf{q}, \mathcal{K}) = -\log p(\mathbf{k}^+|\mathbf{q}) \quad (6)$$

The normalisation constant $Z(\mathbf{q})$ in the denominator of the non-parametric softmax in (5) is expensive to evaluate because it needs to sum over all the negative keys in the dataset for a given query. Noise Contrastive Estimation (NCE) [37], [38] is an estimation method for an unnormalised probabilistic model that avoids the need to evaluate the partition function through a proxy binary classification task, where the binary task is to discriminate between *data samples* (positive keys) and the *noise sample* (negative keys).

Following the original NCE formulation and assuming a uniform *noise distribution* of negative samples $p^-(\cdot|q) = 1/n$ and that we sample noise negative keys $m$ times more frequently than the positive key, the posterior probability of the pair $(\mathbf{q}, \mathbf{k})$ sampled from the positive distribution $p^+(\cdot, \cdot)$ (denoted by $D = 1$) is:

$$p(D=1|\mathbf{q}, \mathbf{k}) = \frac{p(\mathbf{k}^+|\mathbf{q})}{p(\mathbf{k}^+|\mathbf{q}) + m \cdot p(\mathbf{k}^-|\mathbf{q})} \quad (7)$$

With $p(D = 1|\mathbf{q}, \mathbf{k}) = \frac{1}{1+\exp(S(\mathbf{q},\mathbf{k}))}$ parametrised by a sigmoid function with the similarity scoring function $S(\mathbf{q}, \mathbf{k})$, the approximated NCE binary training objective then becomes:

$$\mathcal{L}_{NCE-binary}(\mathbf{q}, \mathcal{K}) = -\mathbb{E}_{p^+}[\log p(D=1|\mathbf{q}, \mathbf{k})] \\ -\mathbb{E}_{p^-}[\log(1 - p(D=1|\mathbf{q}, \mathbf{k}))] \quad (8)$$

This NCE objective has been used widely in learning language models [71] and word embeddings [70]. A slightly different variation of binary NCE is Negative Sampling (NEG) [68] which focuses on learning good word embeddings.

Instead of having a binary task that decides whether each key is positive or negative, suppose we want to correctly identify and rank the positive key with highest similarity to the query in a set $\mathcal{K} = \{\mathbf{k}^+, \mathbf{k}_1^-, \ldots, \mathbf{k}_n^-\}$ with one positive key and $n$ negative keys. Jozefowicz *et al.* [54] extended the *local* view of binary NCE to a *global* or *ranking* view, such that the conditional distribution of key at index $i$ is the positive key is given by:

$$p(i|\mathbf{q}, \mathcal{K}) = \frac{p^+(\mathbf{k}_i|\mathbf{q}) \Pi_{j \neg i} p^-(\mathbf{k}_j|\mathbf{q})}{\sum_{n=1}^N p^+(\mathbf{k}_n|\mathbf{q}) \Pi_{j \neg n} p^-(\mathbf{k}_j|\mathbf{q})} \quad (9)$$

If we let $p(i|\mathbf{q}, \mathcal{K}) = \frac{\exp(S(\mathbf{q},\mathbf{k}_i))}{\sum_j^N \exp(S(\mathbf{q},\mathbf{k}_j))}$ be parameterised by a softmax function, the approximated global ranking NCE training objective then becomes:

$$\mathcal{L}_{NCE-global}(q, \mathcal{K}) = \mathbb{E}_{P(i|q,\mathcal{K})} \left[ -\log \frac{\exp(S(q, k^+))}{\sum_{k \in \mathcal{K}} \exp S(q, k)} \right] \quad (10)$$

The reader is referred to [66], [95] for more detailed treatment of different variations of NCE-based objectives.

Sharing the same motivation with the *Lifted Embedding loss* from the metric learning objective instead of by the NCE objective, Sohn [92] independently proposed the *Multi-class n-pair* loss that has the same formulation as the NCE-global objective in Eq. (10) and uses samples in the same mini-batch as the negative samples to save memory during computation. By formulating it as a multi-class classification problem, this loss automatically incorporates multiple negative keys for comparison, and is thus very effective.

In more recent work, a slightly different form of this loss called the *normalised-temperature cross-entropy (NT-Xent)* [16] loss with a temperature parameter $\tau$ to control sensitivity of the cosine similarity scoring function is used

$$\mathcal{L}_{NT-Xent}(\mathbf{q}, \mathcal{K}) = -\log \frac{\exp(\frac{\mathbf{q}^\top \mathbf{k}^+}{\|\mathbf{q}\|\|\mathbf{k}^+\|\tau})}{\sum_{k \in \mathcal{K}} \exp(\frac{\mathbf{q}^\top \mathbf{k}}{\|\mathbf{q}\|\|\mathbf{k}\|\tau})} \quad (11)$$

The temperature $\tau$ has the same effect of controlling the attraction-repulsion radius around the query, similar to the margin $m$ in the margin-based contrastive loss in Section III-E2.

### 4) MUTUAL INFORMATION-BASED LOSSES

Mutual Information (MI) has a long history in representation learning for various methods that aim to maximise the MI a representation $\mathbf{z}$ and its inputs $\mathbf{x}$. In the same spirit, contrastive learning methods motivated from MI aim to learn a mapping that maximise the mutual information between representations of different views of the same scene, which is upper bounded by the MI between the representation and the input of a scene.

Oord, Li, and Vinyals [77] first proved that minimising the InfoNCE loss based on NCE is equivalent to maximising a lower bound on the MI. Inspired from NCE, InfoNCE comes to the same formulation of the classification-based *N-pair loss* in Eq. (10), and shows that minimising this loss also maximises a lower bound on the mutual information between the input and the representation. Having the same form as the multi-class n-pair loss [92] and NT-Xent [16] but using a bi-linear layer as a scoring function instead of a dot product, this form of contrastive loss is currently the most popular due to its effectiveness and simplicity in implementation, as well as a theoretical guarantee based on MI.

Proposed independently of InfoNCE, DIM [46] also formulated the contrastive learning problem as MI maximisation and evaluated different MI estimators, such as the Donsker-Varadhan (DV) [25], the Jensen-Shannon estimator [74] and the InfoNCE [77].

VOLUME 8, 2020  13

Some recent work [83], [88] performed a review of different MI estimators and derived a new continuum of multi-sample lower bounds that describes the bias-variance and efficiency-accuracy tradeoffs, as well as showing the generalisation bound of MI in the context of contrastive learning.

However, even though mutual information is a principled motivation for contrastive losses based on the information bottleneck principle, simply maximising the mutual information in positive pairs does not guarantee a successful application of the contrastive loss concept. Tschannen *et al.* [102] argue and provide empirical evidences that the success of contrastive losses can not be attributed to mutual information alone.

## IV. DEVELOPMENT OF CONTRASTIVE LEARNING

Now we will briefly examine the major developments in contrastive methods over time, that span over multiple sub-fields and domains.

The core idea of learning by comparing between separate but related data points, without any supervised signal, dates back to 1992 to work by Becker and Hinton [8] and by Bromley *et al.* [11] in 1993. While Becker and Hinton [8] formulate the problem as learning invariant representations by maximizing mutual information among different views of the same scene, Bromley *et al.* [11] introduces the "Siamese Network" composed of two identical weight-sharing networks in a metric learning setup. These are the first examples of the general principle of learning by directly comparing between different training samples.

In 2005, Chopra, Hadsell, and LeCun [20], [39] created the foundation for the contrastive learning framework with the original contrastive pair loss for discriminative models to learn an invariant mapping for recognition and verification problems. Instead of having to define non-linear similarity relationships using some simple metric in the input space, the contrastive pair loss demonstrates the ability to learn a representation space in which a simple distance metric in the embedding space approximates a notion of similarity in the input space.

Inspired by a form of triplet loss used in [108], Collobert and Weston [21] trained an unsupervised language model, and Chechik *et al.* [15] learned an image similarity model using a ranking triplet loss. Later, the triplet loss was applied in the context of a deep neural network and has been shown to be capable of learning fine-grained image similarity [105], or a useful representation [49].

To address the limitations of slow convergence and instability of the pair and triplet contrastive losses, Oh Song *et al.* [75] and Sohn [92] proposed loss functions that improve the number of comparisons for a query in an iteration. While using hard negative and positive samples has been a common component in successfully applying contrastive methods, Wu *et al.* [109] and Hermans, Beyer, and Leibe [45] argue for the case that quality of data pairs used in training are also of paramount importance for pair and triplet losses in the metric learning setting.

While there have been approaches to using probabilistic approaches to learning metric embeddings [98], most successful applications up to now all use the energy-based pair or triplet loss due to the computational requirements to compute the normalisation constant in probabilistic loss. In 2010, Gutmann and Hyvärinen [38] introduced Noise Contrastive Estimation (NCE), a simple conceptual strategy for estimating an unnormalised statistical model by contrasting between the data and noise distributions.

In natural language processing that processes discrete input text tokens, this form of NCE-based contrastive loss has been used to train powerful language models [71] or to learn useful word embeddings [68], [70] from a large unlabelled corpus of text.

Also motivated from the mutual information maximisation perspective similar to [8], in 2018 CPC [77] and DIM [46] made the connection between minimising a contrastive loss with maximising a lower bound of the mutual information between different views.

The instance discrimination task that drove the progress of contrastive methods in the past few years is introduced in [110]. Simplifying the framework for instance discrimination and focusing on learning representations with only augmentation methods, Ye *et al.* [117] and Misra and Maaten [69] showed that pre-training with contrastive loss can outperform supervised-only training for a computer vision task. To achieve the best results with contrastive loss, training with large batch sizes on a large GPU cluster is required. Methods such as Momentum Contrast (MoCo) [43] were introduced to reduce the requirement for large batch sizes. Using an online and momentum-updated offline network, MoCo proposed to view contrastive learning as a form of dictionary lookup and raised questions around how best to retain consistency between offline and online networks to perform similarity matching between the queries and keys.

Using extra network heads on top of the learned representation has been used previously, but it was mostly out of necessity, for example to aggregate context information from multiple time steps such as in CPC [77]. SimCLR [16] proposed an explicit projection head to separate between the tasks of learning a representation and optimising for the contrastive objective. This distinction raises the question of what are the optimal design choices for the base encoder and representations for recent work such as SimCLRv2 [17]. This separation enabled other work to use multiple heads and contrastive objectives when optimising for the same underlying representation [28], [111].

Local aggregation [120] spearheaded the direction of combining clustering methods with instance discrimination contrastive learning, while in [28], [34], [106] the authors raised the question of whether negative samples are necessary at all where they propose a different contrastive loss function to avoid the collapse of the representation with additional implicit constraints.

Table 1 provides a brief summary of some prominent papers over the development of contrastive learning.

TABLE 1. A table summary of the development of contrastive learning methods. Entries are sorted in chronological order of first disclosure. The topics of contribution include *foundational* ideas behind contrastive learning, the development for different forms of the contrastive *loss*, how *similarity* is defined and new *applications* of contrastive learning methods.

| Paper | Short description | Topics of contribution |
|---|---|---|
| Becker and Hinton [8] | Maximise MI between two views | Foundation |
| Bromley et al. [11] | Siamese network in metric learning setting | Foundation |
| Chopra, Hadsell, and LeCun [20] | Learn similarity metric with contrastive pair loss | Energy-based loss, Application |
| Hadsell, Chopra, and LeCun [39] | Learn invariant representation from pair loss | Energy-based loss, Application |
| Weinberger, Blitzer, and Saul [108] | Learn distance metric with triplet loss | Energy-based loss |
| Collobert and Weston [21] | Learn language model with triplet loss | Application |
| Chechik et al. [15] | Learn image retrieval model with triplet loss | Application |
| Noise Contrastive Estimation [38] | Introduce NCE, a general methods to learn unnormalised probabilistic model | Probabilistic loss |
| Mnih and Teh [71] | Learn language model with NCE-based loss | Application |
| Mikolov et al. [68] | Learn word embedding with Negative Sampling (NEG), a modified version of NCE | Probabilistic loss, Application |
| Wang et al. [105] | Learn fine-grained image similarity using deep network and triplet loss | Application |
| Wang and Gupta [107] | Use video's sequential coherence to learn unsupervised video representation | Similarity, Application |
| Lifted-structure loss [75] | Extend triplet loss to multiple positive and negative pairs per query | Energy-based loss |
| N-pair loss [92] | Proposed non-parametric classification loss with multiple negative pairs per query | Probabilistic loss |
| Wu et al. [109] | Focus on the quality of negative samples through a distance-weighted margin loss | Similarity, Energy-based loss |
| Hermans, Beyer, and Leibe [45] | State the important of mining hard samples in triplet loss | Similarity |
| Wu et al. [110] | Self-supervised representation with instance discrimination<br>Memory bank to holds keys for next epoch | Application<br>Encoder |
| CPC [77] | Mutual Information with the contrastive loss<br>Define similarity with past-future context-instance relationship | Mutual Information loss<br>Similarity |
| DIM [46] | Evaluate multiple mutual information bound for the contrastive loss<br>Global-local context-instance relationship | Mutual Information Loss<br>Similarity |
| MoCo [43] | Use momentum encoder to store features to memory queue | Encoder |
| SimCLR [16] | Simplify and demonstrate large empirical improvement in instance discrimination task<br>Focus on the use of separate heads | Application<br><br>Transform heads |
| BYOL [34] | Learning similarity without negative samples | Loss |

## V. APPLICATIONS

We now look at various data domains and problem topics to which contrastive learning representations have been applied, most of this work being very recent. This is done through the lens of the generalised Contrastive Representation Learning framework introduced in Section III.

### A. LANGUAGE

Following the idea proposed in [76] to learn a language model discriminatively, Collobert and Weston [21] learned a language model to perform a two-class classification task to determine whether and how the middle word of a context window is related to its context or not. They used positive examples as instances of such word triples taken from Wikipedia and created negative examples by replacing the middle word in a triplet by a random word and trained the model with a triplet loss.

Later, Mnih and Teh [71] adapted NCE [38] and proposed a more efficient algorithm to learn a language model using a probabilistic contrastive loss, where the context query includes all the previous words, the positive key is the next word in a sequence and the negative keys are sampled from a unigram distribution of words in the corpus.

With the introduction of the Skip-gram and CBOW algorithms [68] to learn word representations which depend heavily on the tree structure of the hierarchical softmax, Mnih and Kavukcuoglu [70] used NCE to avoid having to compute the normalisation term of the softmax. Also inspired by NCE, Mikolov *et al.* [67] proposed a slightly different method called Negative Sampling (NEG) that focuses solely on learning good word representations with the trade-off of losing the probabilistic properties from NCE.

Recently, the Bidirectional Encoder Representation from Transformer (BERT) [24] model learns bidirectional word representations using the Transformer architecture's decoder [103] and demonstrated great performance for transfer learning in multiple downstream tasks. XLNet [116] modified BERT's masked language model objective to include an autoregressive objective. While these language model objectives are usually referred to as a form of denoising autoencoder that try to reconstruct the original input, in the case of learning word embeddings which is just a lookup layer from index to vector, there is no difference between reconstructing and contrasting between feature vectors and thus this work does fall under the remit of being a form of contrastive learning.

Under the mutual information maximisation framework, Kong *et al.* [61] showed that BERT or XLnet also maximise global-local mutual information, whereas the next sentence prediction pre-training task can be seen as constructing similarity pairs using the sequential coherence property. With this insight, Kong *et al.* [61] also proposed BERT-NCE, a variant of BERT that uses an NCE-based loss instead of the full softmax over the entire vocabulary, making it more aligned with contrastive learning methods. Inspired by DIM [46], they also introduce InfoWord that aims to maximise the mutual information between local and global representations of a sentence. The queries for the global representation are the sentence with a contiguous masked chunk which is an n-gram, the positive keys are the local representation of the original n-gram while negative keys are randomly sampled n-grams. The final model used InfoNCE loss to minimise the mutual information lower-bound for both the masked language model and the global-local representation objective.

**TABLE 2.** A summary of methods that applied contrastive methods on language data. The color for defining similarity in query and keys encodes: Multi-sensory, Data transformation, Context-Instance, Sequential Coherence, Clustering. Colors for encoder represent: End-to-end, Online-Offline, Pre-trained. Colors for transform head represent: Projection, Contextualisation and Quantisation.

| Method | Query | Positive keys | Negative keys | Encoder | Transform head(s) | Loss |
|---|---|---|---|---|---|---|
| Collobert and Weston [21] | Surrounding words | Centre word | Random words | Embedding | Max-pooling | Triplet loss |
| Mnih and Teh [71] | Previous words | Next word | From unigram distribution | Bi-linear | Position-dependent weighting | Binary NCE |
| word2vec [70] | Surrounding words | Center word | From unigram distribution | Vector Bi-linear | Position-dependent weighting | Binary NCE |
| word2vec [68] | Surrounding words | Centre word | From modified unigram distribution | Vector Bi-linear | Position-independent weighting | NEG |
| QuickThought [64] | Surrounding sentences | Centre sentence | Sentences outside windows | GRU | No | NEG |
| CPC [77] | Past sentences | Next sentences | Random sentences | ConvNet | GRU | InfoNCE |
| Bert-NCE [18] | Masked sentence | Masked word | Random words | Embedding | Transformer | Binary NCE |
| Sentence-Bert [86] | Query sentence | Same-paragraph sentences | Random sentences | Transformer | Pooling | Triplet loss |
| InfoWord [18] | Sentences with masked n-gram | Original masked n-gram | Random n-gram | Transformer | No | InfoNCE |

In learning representations for units larger than words, Quick-Thought [64] extends the Skip-gram model for word embedding to learn representations for entire sentences. A GRU [19] encodes word-by-word a query sentence and a nearby sentence as the positive keys, while the negative keys are encoded from sentences outside the context window. The final hidden state of the GRU is treated as the sentence embedding.

CPC is a general contrastive learning method that can be applied to many different data modalities. For text data, CPC encodes the context query using past sentences with the positive keys as the future sentence. A 1-D convolution network is used as the encoder to encode the entire sentence, while a GRU acts as a context head and aggregates information from past sentences to predict the representation of future sentences.

SentenceBERT [86] extended word representations from BERT to explicitly learn a sentence embedding using the triplet loss. Two sentences from the same paragraph are considered positive pairs and are negative otherwise. After obtaining individual word representations from BERT, either the special token CLS or a pooling operation is used over the entire sentence to obtain the sentence representation.

Inspired from the success of data transformation-based contrastive methods in computer vision, Fang et al. [29] extended this idea and introduces CERT to learn sentence-level representations. To create positive pairs of sentences, CERT creates two different sentences which are similar in meaning by back-translating, using a machine translation model to translate a sentence into a target language and using another translation model to convert it back to the source language. CERT uses BERT as its encoder and uses InfoNCE as the contrastive loss function.

As yet another alternative approach, Chi et al. [18] used contrastive methods to learn cross-lingual sentence representations using a parallel corpus. In InfoXML, the objective includes a combination of maximising monolingual and cross-lingual token-sequence (global-local) information, and cross-lingual sentence-sentence (multiview) information. The CLS token from the base BERT encoder is used as the sentence representation with a linear projection head. A momentum encoder is used to encode the query while the online encoder is updated using the InfoNCE loss.

Not limited to natural language but still a form of language, [51] learns a functional-equivalent of program code representation by generating similar code snippets using different augmentation techniques from the compiler literature. The transformer's representation of each token is averaged to obtained the representation for the entire program and InfoNCE is used as the contrastive loss.

A summary of the methods that learn language representations using Contrastive learning is shown in Table 2.

### B. VISION
Motivated by the challenges of recognition, verification and fine-grained classification problems, Chopra, Hadsell, and LeCun [20] introduced the contrastive pair loss function in the context of metric learning. Such applications need to deal with data with high intra-class variance (e.g same face but different lighting condition and angles) and low inter-class variance (e.g different faces but taken by the same camera setup). The explicit formulation of a contrastive learning objective to minimise the distance between inputs of the same class whilst maximising the distance between inputs of different classes is a direct attempt to solve this problem. On the other hand, Hadsell, Chopra, and LeCun [39] demonstrated that the contrastive loss will learn an invariant mapping for many irrelevant input features in order to be able to map different inputs to the same neighbourhood in the embedding space.

Building on the intuition of invariant mapping and its application in metric learning, Chechik *et al.* [15] learned a large scale image similarity model for retrieval using the triplet loss.

Moving beyond metric learning applications, Hoffer and Ailon [49] used a similar triplet architecture but focused on learning image representations simply from using the class labels to denote similar pairs. Wang and Gupta [107] extended this idea beyond supervised learning by learning visual representations from video with the help of an unsupervised tracking method. The corresponding patches provided by the tracker are used as the positive pairs while the hard negative pairs are mined from elsewhere in the dataset.

Among the first to exploit sequential coherence for defining triplets, Sermanet *et al.* [91] introduced the Time-Contrastive Network (TCN), a self-supervised method to learn a view-agnostic but time-sensitive representation from unlabelled videos. Two simultaneous views from different cameras, or two consecutive frames from the same view are defined to be similar, while two frames far apart in time but from the same camera view are defined to be dissimilar.

Recently contrastive learning has received a lot of attention due to its successful application to self-supervised visual representation learning, especially in the Instance Discrimination task introduced by Wu *et al.* [110]. Following the idea of treating each instance as its own exemplar class [26], a memory bank mechanism was introduced to store the computed representations for use in future iterations, so that the number of negative samples is decoupled from the batch size. The queries are computed online and contrasted with the keys from the memory bank where the global NCE objective is used to learn to discriminate between features of the same instance or not. Looking at contrastive learning as a dictionary lookup problem, He *et al.* [43] introduce Momentum Contrast that maintains the offline encoder as an exponentially weighted average of the online encoder where it stores the key representations in a queue, weighting more recent key representations as being more important.

Since the difference between the query and the positive keys in instance discrimination is how they are randomly augmented, multiple works such as *Invariant and Spreading Instance Feature* [117], PIRL [69], SimCLR [16] have focused on engineering strong and varied augmentations to yield better representation from the ImageNet [23] dataset without class labels. These methods have attracted special interest because for the first time they outperform supervised ImageNet classification pre-training on multiple downstream vision tasks. SimCLRv2 [17] performed a comprehensive study of contrastive self-supervised learning in semi-supervised settings where few labels are present, and demonstrated state-of-the-art results by contrastive pre-training in various downstream vision tasks.

In a different direction, Oord, Li, and Vinyals [77] proposed CPC to learn invariances between context-instance relationships instead. The predictive coding principle in CPC defines context as the past, and that a good representation of the past will possess a strong predictive capability for instances in the future. The predictive power of a representation is modelled as a contrastive objective that maximises the mutual information between the past context and the future instance through the InfoNCE mutual information lower bound. While the CPC method is general and equally well applicable to multiple data modalities, CPCv2 [44] improved on CPC with some architectural design changes specifically for learning from images and evaluating this on label-efficient fine tuning tasks. Expanding CPC into learning representations from natural videos, *Dense Predictive Coding* (DPC) [40] contrasts between local patches of the feature maps extracted from the past context with the local patches of the features maps extracted from future instances. DPC employs three kinds of negative samples: the easy negatives come from patches encoded from different videos, the spatial negatives come from the same video but at different spatial locations of the feature maps, and the hard negatives come from the same spatial location but from different time indexes.

Also learning invariances from context-instance relationship, DIM [46] defined context to be a little more general than CPC. A single vector for each image is used as the global representation, while the feature vectors at each spatial location from the feature map at previous layers are considered local features. DIM enforces the contrastive objective using multiple different mutual information lower-bounds but also found that InfoNCE is the most effective, especially with a large number of negative samples. Combining the context-instance strategy with the temporal coherence property of a video, Anand *et al.* [2] proposed *SpatioTemporal DeepInfoMax (ST-DIM)* that learns to maximise mutual information between global features of the current frame and local features from the next frames. Finally, *Augmented Multiscale DIM* [5] combined both the global-local objective from DIM [46] and image data augmentation from the instance discrimination task to learn visual representations.

By exploiting temporal consistency as a natural source of image transformation, *Video Noise Contrastive Estimation* (VINCE) [33] modified the instance discrimination task where instead of contrasting between two augmented views of the same image, VINCE defined positive pairs as two frames from the same video. An additional benefit of this approach is that different objects that are likely to show up in the same video (e.g dog and cat) are also encouraged to be closer than more random pairs (e.g cat and whale).

By combining the image data transformation, temporal coherence between frames and global-local correspondence between features, *Video Deep InfoMax* (VDIM) [47] learned effective spatio-temporal representations for downstream tasks on videos.

Exploiting visual similarity to form natural clusters in the representation space has been used previously to learn unsupervised representations [13]. This objective has been reformulated in the form of a contrastive learning method in [120], where a set of close neighbours is aggregated together from a set of background neighbours. Given a query image, the *background neighbours* are an unbiased sample of nearby points measured with cosine distance in the embedding space. An unsupervised clustering algorithm is applied on the set of background neighbours, where the samples in the cluster that includes the query are the *close neighbours*, which act as the set of positive samples for that query. The embedding is learned iteratively using an NCE loss to classify between close neighbours and background neighbors. In addition to just preserving the local smoothness around each instance in the same cluster, *Prototypical Contrastive Learning (PCL)* [63] also encoded the higher semantic structure of the data into the embedding through the cluster's centroid. Assuming that each data point is associated with a latent class variable, PCL aims to learn both the class's prototype and optimises for points belonging to a cluster to stay close together through the Expectation Maximisation (EM) framework. In the E-step, k-clusters are obtained by performing *k*-means on the features from the momentum encoder and the distance from each point to its cluster's prototype is minimised using the InfoNCE loss in the M-step.

Most clustering-based methods up to now are offline in the sense that they require multiple passes over the data to compute features and perform clustering, but *Swapping Assignment between multiple Views (SwAV)* [14] proposed an online clustering method to learn unsupervised visual representations. Combined with data transformation approaches in instance learning, two different augmented views of the same images are encoded into features and the clustering assignment for each of the views is computed from a set of trainable "code" vectors. Similarity is enforced through a "swapped" prediction problem where the feature vectors from one of the views is matched with the cluster's code from the other views. No negative pairs are explicitly used in this method but the representation is prevented from collapsing through the batch-wise online code computations. *InterCLR* [112] also performed mini-batch clustering with a set of learned cluster centroids but instead of using a swapped prediction with no explicit negative samples, they modelled the instance-cluster relationship by assigning a pseudo-label for each instance. Samples that shared pseudo-labels are positive pairs while samples that have different labels are negative pairs. All of these clustering-based contrastive methods in a sense enhance the similarity and dissimilarity in the instance discrimination task through using pseudo-labels derived from clustering techniques.

Most of the methods above focus on the self-supervised paradigm and thus refrain from using human-annotated labels. *Supervised Contrastive Learning* [56] directly used class labels to define similarity, where samples from the same class are positive and samples from different classes are negative samples. This method was shown to be more robust to corruption than using the usual cross-entropy loss with the labels alone.

Most of the work above utilised the NCE objective in one form or another, which will usually benefit with more negative samples. Therefore self-supervised contrastive representation learning methods usually require large batch sizes and longer training times than other supervised or self-supervised methods. The training dynamic of contrastive methods can be dissected into two keys properties [106], *alignment* (closeness) of features from positive pairs and *uniformity* (spreading) of the induced representation on a hypersphere. The *uniformity* explains the role of negative pairs in keeping the representation from collapsing and opens up the research direction of using other methods without negative samples to prevent the representation from collapsing. In *SwAV*, similarity is formulated as a swapped prediction problem between positive pairs while the minibatch clustering methods implicitly prevent collapse of the representation space by encouraging samples in a batch to be distributed evenly to different clusters. In *Bootstrap Your Own Latent (BYOL)* [34], the similarity constraint between different views are also enforced through a prediction problem, but from an online network to an offline momentum-updated network. The key insight is that by trying to match the prediction from an online network to a randomly initialised network, the obtained representations are already better than those of the random offline network. By continually improving the offline network through the momentum update, the quality of the representation is bootstrapped from just the random initialised network.

In concurrent work, Ermolov *et al.* [28] proposed a *Whitening MSE* loss, where again the similarity between augmented instances is enforced through the minimisation of MSE distance in the embedding space, while the *whitening* operation common in many image pre-processing pipelines is applied on the representation in batch. The whitened vectors of all samples in a batch, including positive pairs, become distributed and the MSE objective will pull features of positive pairs closer together i.e. the distance between positive pairs is small while the representation space does not collapse into a single cluster.

A summary of the methods that learn visual representations using Contrastive learning is shown in Table 3.

### C. AUDIO

For audio processing, CPC [77] used a strided convolutional network as the base encoder to map from raw audio signal to the representation **v** where a GRU RNN head aggregates the information from all previous timesteps to form a contextualised representation **z**. This contextualised embedding **z** is then used as the query where it is contrasted with a set of representations $\mathbf{v}$ with respect to the true future $\mathbf{v}^+$ from the noise $\mathbf{v}^-$.

**TABLE 3.** A summary of methods that applied contrastive methods on vision data. The color for defining similarity in query and keys encodes: Multi-sensory, Data transformation, Context-Instance, Sequential Coherence, Clustering. Colors for encoder represent: End-to-end, Online-Offline, Pre-trained. Colors for transform head represent: Projection, Contextualisation and Quantisation.

| Method | Query | Positive keys | Negative keys | Encoder | Transform head(s) | Loss |
|---|---|---|---|---|---|---|
| Chechik et al. [15] | Query image | Same label images | Different labels images | Bag-of-local-descriptors | None | Triplet loss |
| Hoffer and Ailon [49] | Query image | Same label images | Random images | Multi-scale ConvNet | None | Triplet loss |
| Wang and Gupta [107] | Patch from first frame | Tracked patch from last frame | Random sampling and hard negative mining | ConvNet | None | Triplet loss with cosine distance |
| TCN [91] | Frame $t$ from camera 1 | Frame $t$ from camera 2 | Frame $t+k$ from camera 1 | ConvNet | None | Triplet loss |
| Wu et al. [110] | Augmented image | Same image from memory bank | Random images from memory bank | Convolutional Net + Memory Bank | None | Non-parametric classification |
| MoCo [43] | Augmented image | Augmented query image | Random images from momentum queue | Convolutional Net + Momentum Encoder | Linear | InfoNCE |
| SimCLR [16] | Augmented image | Augmented query image | Random images from batch | Convolutional Net | MLP | NT-Xent |
| CPC [77] | Aggregated patch's features | Subsequent patches | Random patches | Convolution | Convolutional row-GRU | InfoNCE |
| DIM [46] | Global feature | Local feature maps of query | Local feature maps of random images | Convolution Net | Convolution Net + Pooling | InfoNCE |
| ST-DIM [2] | Global feature at time step $t$ | Local feature map at time step $t+1$ | Local feature map at random time step $t^*$ | Convolution Net | MLP | InfoNCE |
| AMDIM [5] | Augmented global feature | Augmented multi-scale local features of query | Multi-scale feature map of random images | Convolution Net | Convolutional Net + Pooling | InfoNCE |
| VINCE [33] | Query frame | Frames from same video | Frames from random videos | Convolutional Net + Momentum Encoder | MLP | InfoNCE (multiple positive pairs) |
| VDIM [47] | Global features of a frame sequence | Global and local features of subsequent frames | Global and local features of random frame sequences | Video Convolutional Net | Convolution Net | InfoNCE |
| Local Aggregation [120] | Query image | Close neighbours | Background neighbours | Convolutional Net + Memory bank | Linear | InfoNCE |
| PCL [63] | Augmented image | Augmented and prototypes vectors of query | Feature and prototypes vectors of random images | Convolutional Net + Momentum encoder | MLP | ProtoNCE (instances + clusters InfoNCE) |
| SwAV [14] | Prototype of query | Prototype of augmented query | No negative samples | ResNet | MLP + soft cluster assignment | Instances + clusters InfoNCE |
| InterCLR [112] | Augmented query image | Images with same cluster's pseudo-label | Images with different cluster's pseudo-label | Convolutional Net + Memory bank | MLP | InfoNCE |
| Khosla et al. [56] | Augmented image | Images with same supervised label | Images with different supervised label | ConvNet | MLP | InfoNCE (multiple positive pairs) |
| BYOL [34] | Augmented image | Augmented query image | No negative pair | Convolutional Net + Momemtum Encoder | Projection MLP + Prediction MLP | MSE |

Built on top of CPC, wav2vec [89] uses another convolutional network to aggregate context information instead of using a recurrent network for the context head. Moving beyond evaluating on frame-wise phoneme classification in CPC, Schneider *et al.* [89] evaluated the learned representation of wav2vec and applied the contrastive pre-trained representation to improve a supervised Automatic Speech Recognition (ASR) system. VQ-wav2vec (Vector-quantised wav2vec) [6] modifies the wav2vec architecture by using an additional quantisation head before the context head.

The quantisation head is implemented through a Gumbel-softmax [52] to convert the continuous speech signal $\mathbf{v}$ into a set of discrete codes $\mathbf{c}$. The context head is built on top of these discrete codes to form the query context vector $\mathbf{z}$. Similar to CPC and wav2vec, the context vector is then compared with another quantised representation $\mathbf{c}$ to find the representation of the correct future. The discretised speech representation can then be used directly as a representation for other models that expect discrete input such as BERT [24].

All of these methods above encode context representation using only past-to-present information. Inspired from the success of the bidirectional encoding in the transformer model [103], Wav2vec 2.0 [7] replaces the unidirectional context head from vq-wav2vec [6] with a bidirectional masked Transformer.

**TABLE 4.** A summary of methods that applied contrastive methods on audio data. The color for defining similarity in query and keys encodes: Multi-sensory, Data transformation, Context-Instance, Sequential Coherence, Clustering. Colors for encoder represent: End-to-end, Online-Offline, Pre-trained. Colors for transform head represent: Projection, Contextualisation and Quantisation.

| Method | Query | Positive keys | Negative keys | Encoder | Transform head(s) | Loss |
|---|---|---|---|---|---|---|
| CPC [77] | Aggregated past context | Future signal | Random signal from same audio clip | Convolution | GRU | InfoNCE |
| Wav2vec [89] | Aggregated past context | Future signal | Random signal from same audio clip | Convolution | Convolution | InfoNCE |
| VQ-Wav2vec [6] | Aggregated past context | Future signal | Random signal from same audio clip | Convolution | Convolution + Gumbel softmax | InfoNCE |
| Wav2vec 2.0 [7] | Masked bidirectional vector | Masked quantised vectors | Random quantised vectors from same clip | Convolution | Gumbel softmax + Masked Transformer | InfoNCE |
| Nandan and Vepa [73] | Augmented mel spectrogram | Augmented mel spectrogram | Random Mel spectrograms | Convolution | MLP | InfoNCE |

In a different direction, Nandan and Vepa [73] learned speech representation from audio in mel spectrogram image format. Combined with mel spectrogram data transformation techniques (i.e time and frequency masking [79]), they use a pipeline similar to many image instance discrimination methods to a learned representation that is language agnostic and is shown to transfer well to an emotion classification task, regardless of the spoken language.

A summary of the methods that learn an audio representation using Contrastive learning can be seen in Table 4.

### D. GRAPHS

For relational and graph-structured data, contrastive learning has been successfully applied to learn both node, edge and graph-level representations.

The earliest approaches to learning representation from relational data that comes in the form of triplets *(subject, relation, object)* is Linear Relational Encoding (LRE) [78]. In this early work, the representation encoder is just a simple embedding layer for the *subjects* and *objects*, while the *relations* are represented as a matrix. The transform head in this case is a simple matrix-vector multiplication between the *relation* and *subject*, so that the resulting vector is closest to that of the *object*.

Later, Bordes *et al.* [10] introduced TransE, which learns a vector embedding for both the nodes and edges, and uses an additive transform head to represent relations as a translation in the embedding space. TransE uses an energy-based triplet loss to learn the embeddings and similar to LRE, the negative training pairs are created by corrupting the *object* node with random nodes from the data.

More recently, the Contrastively-trained Structured World Model (C-SWM) [60] uses a Graph Neural Network to model each state embedding as a set of objects and their relations. The base encoders consist of a CNN object extractor and an MLP object encoder, that turn an image into an abstract state representation. The graph Neural network heads then transform the state's representations and its corresponding actions (represented as one-hot vectors) into the state representation in the next time step. Similar to TransE, the state transitions between time steps is modeled as a translation in the embedding space and the entire world model is trained end-to-end with an energy-based hinge loss.

Focusing on learning useful node representations from general graphs, node2vec [35] aims to learn a node representation that is similar between neighbour nodes. The key contribution of node2vec is a family of biased random walk methods, allowing for a flexible notion of network neighbourhood (i.e positive keys). The model is trained similar to the Skip-gram model in word2vec, using negative sampling.

Veličković *et al.* [104] follows DIM [46] to propose Deep Graph Infomax (DGI) to learn node embedding by maximising mutual information between representations of local and global patches of a graph. The encoder is a Graph Convolutional network [31], [59] that summarises a patch of the graph centered around some nodes. A contextualisation head in the form of a *readout function* summarises the patch representations into a graph-level global representation so that all patches encode the most useful features present in the global features. The negative samples are patches from random graphs in a multi-graph setting or a corrupt function is used in a single-graph setting.

Also inspired by the mutual information maximisation between global and local structure of DIM, but with some design choices different from DGI [104], InfoGraph [97] focuses on learning graph-level representations. InfoGraph uses GIN [115] as the base encoder and uses sum over mean for the readout function, both of which are more suitable to learning representations at graph-level.

Combining both the multi-view and global-local mutual information maximisation objective, Hassani and Khasahmadi [41] aims to learn both graph-level and patch-level representations for graphs. A graph diffusion is used to generate a different structural view of the graph, and then a sub-graph is sampled from both of the views. A dedicated GNN is used as the base encoder for each view, while the transform heads are shared between the two views. An MLP is used as projection head for the node representation, while a pooling layer followed by an MLP is used as the contextualisation head for the graph representation. A mutual information contrastive loss is then used to maximise the similarity between a local representation of one view to a global representation of another view.

**TABLE 5.** A summary of methods that applied contrastive methods on relational and graph-structured data. The color for defining similarity in query and keys encodes: Multi-sensory, Data transformation, Context-Instance, Sequential Coherence, Clustering. Colors for encoder represent: End-to-end, Online-Offline, Pre-trained. Colors for transform head represent: Projection, Contextualisation and Quantisation.

| Method | Query | Positive keys | Negative keys | Encoder | Transform head(s) | Loss |
|---|---|---|---|---|---|---|
| LRE [78] | Concept + Relation | Paired Concept | Random concept | Embedding | Multiplication | Probabilistic loss |
| TransE [10] | Concept + Relation | Paired Concept | Random concept | Embedding | Addition | Triplet loss |
| Node2vec [35] | Query node | Neighbour node | Random node | Embedding | No | NEG |
| DGI [104] | Global graph | Local graph | Corrupted local graph | GCN | Readout average | Binary NCE |
| C-SWM [60] | State + Action | Next state | Random state | CNN and GCN | Addition | Pair loss |
| InfoGraph [97] | Global graph | Local graph | Random local graph | GIN | Readout summation | JSD MI estimator |
| Hassani and Khasahmadi [41] | Global graph | Transformed local graph | Random local graph | Graph ConvNets | Readout + Pooling for global, non-linear for local graph | JSD MI estimator |
| GCC [85] | Sub-graph structure | Transformed sub-graph | Random sub-graph | Momentum GIN | No | InfoNCE |

Aiming to learn a structural representation of a graph without node attributes and labels, Graph Contrastive Coding (GCC) [85] simulates the augmentation-based instance discrimination task in computer vision. GCC treats each sub-graph as an instance and tries to learn a representation that captures similarity between sub-graphs by discriminating between these instances. A positive key is created by applying a *graph sampling* transformation on that sub-graph. GIN [115] is used as the base encoder with a momentum encoder [43] for the keys and InfoNCE is used as the contrastive loss.

A summary of the methods that learn graph representations using Contrastive learning is shown in Table 5.

### E. MULTI-MODAL

The constraints enforced by the contrastive loss distance metric are not limited to embeddings from the same media modality. Contrastive learning has also been used to to learn cross-modal embeddings from two or more modalities that enhances the representation learned from a single data modality, especially for data that has limited labels.

In the most obvious way, the "views" from *Contrastive Multiview Coding (CMC)* [99] is straightforward to extend to multiple modalities. In this paper, they experimented with views from *L* and *ab* channels from RGB color images, or from one RGB frame and an optical flow feature at the same time.

The *Audio-Visual Correspondence* task is one example where it is desirable to have a joint representation space between representations extracted from the visual and audio modalities. The *Audio-Visual Embedding Network (AVE-Net)* [4] is an example where contrastive learning is applied to this problem. Two separate convolutional encoders for the vision and audio data streams are used. The audio which is 1 second in duration and is centered around the selected frame, is considered a positive pair, while negative pairs are extracted from different videos. This is different from the verification setting from previous work [3], where an MLP fusion network takes the concatenation of the two representations and outputs the final decision on whether the signals correspond.

Instead, AVE-Net explicitly projects representations from each sub-network to a common embedding space through the use of a non-linear MLP head and measures correspondence through a contrastive loss using Euclidean distance in the embedding space. Since similarity between representations is explicitly enforced instead of implicitly learned in the fusion network as in [3], the embeddings learned by AVE-Net [4] are well-aligned and more suitable for cross-modal retrieval tasks.

Similarly, *Cross-modal Audio Visual Instance Discrimination (Cross-AVID)* [72] jointly learn the general representation from video using corresponding image frames and audio segments. In addition to contrasting between audio and visual representations of the same instance, they introduced a *Cross-modal Agreement (CMA)*, a mining method that extends the set of positive pairs beyond just from a single instance. CMA measured the agreement of two videos based on both their visual and acoustic characteristics and if two videos have high agreement in both modalities, they are considered positive pairs.

Performing *within-modal* contrastive learning beyond the instance-level using the extended definition of positive pairs from CMA helps to improve the performance of Cross-AVID, and reduces the chances of the representation collapse phenomenon observed in cross-modal learning settings. Very similarly, Patrick *et al.* [81] performed visual audio cross-modal contrastive learning with a more principled approach to sampling and augmentation in an attempt to qualitatively measure the invariance and covariance, which they refer to as "distinctiveness", captured by the learned embedding.

Instead of contrasting cross-modal representations of different instances, Afouras *et al.* [1] used de-synchronisation to select negative samples by mis-aligning (shifting) the video and audio features. The global features from the audio signal for a frame is compared with the local features from the feature map of the vision network, resulting in an audio-visual attention map. A max-pooling layer acts as the context head to summarise the agreement between the audio and visual signals.

**TABLE 6.** A summary of methods that applied contrastive methods on multimodal data. The color for defining similarity in query and keys encodes: Multi-sensory, Data transformation, Context-Instance, Sequential Coherence, Clustering. Colors for encoder represent: End-to-end, Online-Offline, Pre-trained. Colors for transform head represent: Projection, Contextualisation and Quantisation.

| Method | Query | Positive keys | Negative keys | Encoder | Transform head(s) | Loss |
|---|---|---|---|---|---|---|
| CMC [99] | $L$-channel | $ab$-channel | Random $ab$-channel | Convolutional Net | None | InfoNCE |
| AVE-Net [4] | Query frame | Audio clip centered around query | Random audio clips | Convolutional Net | None | Euclidean distance + linear classifier |
| Cross-AVID [72] | Query video clip | Paired audio clip | Random audio clips | Convolutional Nets | MLP | Binary NCE |
| Patrick et al. [81] | Augmented query video clip | Augmented paired audio clip | Random audio clips | Convolutional Nets | MLP | InfoNCE |
| Afouras et al. [1] | Local features map of video | Global feature of aligned audio | Global feature of misaligned audio | Convolution Nets | Spatial Max-pooling | InfoNCE |
| Jiao et al. [53] | Query video frames | Aligned audio | Misaligned audio | Convolutional Nets | MLP | InfoNCE |
| Sun et al. [96] | Query video | Paired ASR text | Random ASR text | Video: 3D Convolutional Net + Transformer, Text: BERT | Transformer | InfoNCE |
| Ilharco et al. [50] | Object image | Paired text description | Random text description | Text: BERT, Image: Faster RCNN | LSTM + linear | InfoNCE |
| COALA [30] | Query audio | Paired tags | Random tags | Audio: ConvNet, Multi-hot tags: MLP | Non-linear | InfoNCE |
| CSTNet [57] | English speech | Paired translation text | Semi-hard mining translation text | Audio: ConvNet, Text: Word embedding + ConvNet | None | Triplet loss |

Jiao *et al.* [53] applied the misalignment objective to learn joint embeddings for ultrasound audio and the corresponding doctor's narrative speech. Applying contrastive learning in this setting is particularly helpful because this type of paired data is a lot easier to collect in a medical setting. Positive and negative pairs are defined based on spectrum of misalignment in time. Positive and ''hard-positive'' pairs are video frames and their corresponding or slightly misaligned audio clips. Negative and ''hard-negative'' are pairs of frames and audio clips that are even further misaligned from each other in time.

Instead of learning the correspondence directly between the visual and audio signals, in [96] video representations are learned by contrasting with representations from text captions extracted from an Automated Speech Recognition (ASR) system. The ASR sequences are encoded using a pre-trained BERT [24] model while a pre-trained S3D [113] model is used to extract visual features which are then fed into a shallow Transformer [103] network to construct a video-level visual embedding. The scoring function comprises another shallow transformer module that acts on the concatenated representations from the two modalities, followed by an MLP network that estimates the mutual information (MI) between the two inputs. The MI scores between them are again estimated through a softmax classification setting.

Not limited to jointly learning an embedding space, contrastive methods can also be used to learn a mapping between two separately-trained models of different modalities. Ilharco *et al.* [50] learned a probe to find the similarities between words and object images from a paired image captioning dataset. Even though the BERT [24] text encoder and the Faster RCNN [87] object detection model are trained separately and not updated by the contrastive loss, the LSTM cells [48] and a linear project head can still map between words and object representations.

In the same spirit of learning representations from loosely aligned data, *COALA* [30] learns a shared embedding between audio and its tags, which are more readily available than a corresponding audio-transcript. In a different setting, Khurana, Laurent, and Glass [57] demonstrated a proof-of-concept approach to learn a translation network between English speech and its text translations in other languages. Their *CSTNet* used a triplet loss with a semi-hard negative mining method to learn both a cross-modal and cross-lingual representations.

A summary of the methods that learn multi- or cross-modal representation using Contrastive learning is shown in Table 6.

### F. OTHERS

We conclude this section on applications of contrastive learning by looking at some others works that apply contrastive learning on other field such as reinforcement learning or that are different from the usual pre-train then transfer of contrastive representation learning framework in other modalities.

Not limited to learning representations, contrastive learning can also be applied to distill knowledge from a large pre-trained teacher network to a smaller student network, as demonstrated in *Contrastive Representation Distillation (CRD)* [100].

In addition to learning representations of observations in the environment, *CPC|Action* [36] is a variant of CPC that explored whether contrastive learning methods can also encode *belief states* (i.e its uncertainty) in its representation condition on the future action.

To improve the representation for reinforcement learning (RL) tasks, *CURL* [93] applied the instance discrimination task with a momentum encoder from MoCo [43] to train model-free RL agents directly from the pixel observations. Due to the fact that many RL algorithms operate on a sequence of frames, the augmentations to create positive pairs are applied consistently across a consecutive frame stack as opposed to a single frame.

In an attempt to decouple representation and reinforcement learning, Stooke *et al.* [94] proposed the *Augmented Temporal Contrast (ACT)* for pre-training representations that are transferable to multiple RL tasks. Using the temporal consistency properties and a momentum encoder, augmented observations are contrasted with future observations in the same trajectory using the InfoNCE loss.

In a different vision application, Park *et al.* [80] proposed multi-layered patch-wise contrastive methods to enhance the performance of an unpaired image-to-image translation model. With the intuition that for a given patch of a style-transformed image, the corresponding patch at the same layer and spatial location should be more strongly associated with that patch than at any other patches at different spatial locations, InfoNCE contrastive loss is used to maximise the mutual information between patches at the same spatial location of both input and output images.

In other lines of work that try to learn representations in a greedy layer-wise manner instead of through an end-to-end approach using gradient descent, it has been shown that mutual information maximisation through the contrastive InfoNCE loss is particularly suitable for greedy optimisation. In this direction, *Greedy InfoMax (GIM)* [65] extends the approach of CPC [77] while *Local Contrastive (LoCo)* [114] improved the performance by extending SimCLR [16] with a modified overlapping architecture between local layers.

## VI. DISCUSSION AND OUTLOOK

In this section we analyse and raise some questions about the current limitations and possible future directions for contrastive representation learning.

### A. WHAT KIND OF REPRESENTATIONS ARE LEARNED BY CONTRASTIVE METHODS?

Recent successes in transfer learning by instance discrimination contrastive pre-training [16], [43], [69] have raised the question of "what representation is learned from contrastive methods and why is it better than supervised pre-training" [101], [119]? However from the view of the Contrastive Representation Learning framework, the invariant and covariant features learned from the instance discrimination task are entirely decided by the augmentations techniques that create the positive pairs. To understand the effect of augmentations on the representation, one must take into account the bias of the dataset that it was applied to as well. As analysed in [84], models trained with an instance discrimination objective rely heavily on the occlusion invariance property, which was induced by applying aggressive cropping on centred, single-object images from ImageNet [23]. Naively applying this "overfitted" set of augmentations on a different dataset with a more diverse composition of scenes can lead to unexpected behaviour in the representation. To successfully apply contrastive learning to other data sets and problems, one must be aware of the bias represented in the data together with the principle behind how positive and negative samples are produced (Section III-B).

### B. CONTRASTIVE LOSS NEEDS MORE OR NO NEGATIVE SAMPLES?

Based on the theoretical guarantee of NCE and empirical evidence, the performance of contrastive learning methods benefit from comparison with multiple negative samples, which requires training on large GPU clusters and longer training times. One approach to alleviate this problem is to employ memory tricks such as the momentum encoder technique (Section III-C) that can allow the incorporation of even more negative samples and is not limited to the batch size limited by hardware memory. Based on the assumption that negative samples are present just to prevent the representation from collapsing into one single cluster, another direction is to eliminate the need for negative samples altogether and impose additional constraints on the embedding space to prevent it from collapsing [28], [34].

Beside quantities, qualities of negative samples are often neglected as sampled uniformly from the data distribution. More careful selection of negative samples has been shown to improve the convergence rate and performance of the learned embeddings on downstream tasks. This is consistent with hard negative and positive mining techniques, which has been a standard component in many metric learning applications.

This raises the question of a quality vs. quantity trade-off in employing negative samples for contrastive loss. Would it be possible to design a contrastive loss that employs both architectural constraints, perhaps for early stages of learning, and uses hard negative samples to learn a more fine-grained representation in the latter stage?

### C. WHAT AND HOW DO DIFFERENT ARCHITECTURAL DESIGNS AFFECT THE PERFORMANCE OF CONTRASTIVE METHODS?

The separation between the transform heads and base encoder serves as a conceptual distinction to focus on transfer learning on downstream tasks, but in practice the distinction is not so clear cut. While the base encoders are mostly borrowed directly from supervised learning, with some modifications such as wider layers to capture more features, the best choices for projection and transform heads is unclear. In some cases the transform head is necessary (e.g to perform feature aggregation as shown in Section III-B3). Other possible choices are to not use any head, or to use a linear layer and non-linear multi-layers projection heads. In SimCLRv2 [17], empirical experiments show that the output of the second layer of a 3-layer MLP projection head is a better representation for transfer learning than the output of the base ResNet [42] encoder. In BYOL [34], in addition to the projection head from a high-dimensional representation embedding to a lower-dimensional metric embedding, a MLP ''prediction network'' projects metric embeddings of the online to that of the offline networks. This additional bridge between two embedding spaces is a crucial component for the success of the entire model.

These design choices are usually the result of empirical experiments specific to the architecture. The observations suggest a potential discrepancy in architectural design for supervised learning and representation pre-training, as well as potential for research in principles to design an efficient architecture for contrastive methods and representation learning in general.

Another under-explored topic is the specific form of the representation, which is currently treated as a simple vector for each input. With the ease of specifying invariant and covariant properties allowed by the contrastive framework, LooC [111] is an example where contrastive learning is used to concurrently learn multiple embedding sub-spaces, each of which is invariant to all but one transformation as specified by the distribution of positive pairs. Learning disentangled and compositional representations using contrastive learning is a promising research direction.

### D. AN ASYMETRIC SCORING FUNCTION?

Even though the learned similarity score has previously been used for retrieval and ranking applications, currently computing similarity or distance in contrastive learning is mostly used as a proxy task to learn representation. Can the learned similarity score be used in novel applications that were not previously possible?

An interesting possible extension for the scoring function is an asymetric one. The current literature on contrastive methods assume a simple symmetric distance/similarity relationship, but not all kinds of similarity are the same (for example the similarity between ''dog-cat'' should be different from ''dog-animal''). Could a contrastive loss with non-transitive similarity relationships be developed?

### E. FUTURE OF THE CONTRASTIVE LOSS FUNCTION?

As discussed in Section III-E, the form of the contrastive loss is generally motivated from an energy-based margin loss, NCE-based classification or mutual information maximisation. The most popular form of contrastive loss belongs to the family of InfoNCE (and its variants such as NT-Xent), due to its efficiency and simplicity, with a well-grounded motivation from information theory. Can we design better contrastive loss functions that are more efficient in computation and memory, for example one that is more suitable to incorporate multiple positive keys for one query?

From which perspective can such a loss be developed? Even though contrastive losses motivated from mutual information have a strong body of theoretical support, as pointed out in [102], maximising mutual information alone can not explain for all the successes of contrastive learning methods. Looking at the contrastive loss from all the different perspectives may motivate the development of a new generation of contrastive losses.

### F. BEYOND LEARNING REPRESENTATION WITH CONTRASTIVE METHODS

While this paper focuses on the majority of work that applied contrastive learning to learn representation, either supervised or self-supervised, the question of whether learning representation first is actually *necessary*, is still not settled. Even though there is ample evidence that representation learning on a general data stream benefits the performance of models when fine tuned on low-resource tasks, one can argue that if we know the task we want to be good at there are better ways to directly optimise for that task without explicitly dealing with the representation as a leaky abstraction [12]. Because contrastive learning only needs a definition of positive and negative distribution for pairs of samples, one can potentially define those just once for the entire data set or data stream, and optimise directly for a relevant task using the contrastive loss. Therefore contrastive methods can potentially extend task-based learning beyond the need for a static labelled dataset, as is the case for current supervised learning methods.

## VII. CONCLUSION

Although there has been a recent surge of interest in the topic, contrastive learning and contrastively learning representations is not a new idea, with work dating back nearly 30 years to the early 1990s. This is partly because much of the machine learning field is now taken up by problems of data architecture and systems engineering and scalability. This usually involves building systems which are bigger and operating under the maxim that bigger is better. Contrastive learning is more like data engineering and it allows the properties of data to emerge naturally based on data similarity rather than trying to fit data processing into some large and complex system architecture.

In this paper, we introduced a framework for describing contrastive learning and we presented a taxonomy of work done in the field. Because contrastive learning has been used in multiple applications and input domains including image, video, text, audio and others, we have had to draw together input from NLP, computer vision, audio processing and more in order to present a comprehensive survey of the field, with inputs also drawn from across these disparate application areas.

While the paper will provide a useful resource for those who have little background in the topic of contrastive learning and who want to learn more, it will also be of value to those already familiar with the topic since contributions to the development of the area are drawn from such a range of sources.

Contrastive learning and contrastive representations of data represent an interesting and different approach to modeling data which is suited to some kinds of datasets, and for applications where labelled training data may not be available or in sufficient amounts to support typical deep learning approaches.

Whilst successful contrastive representation learning typically involves using relatively more computational resources (and thus power), the models produced by this process often enable rich general-purpose representations that show greater performance on a variety of downstream tasks than their end-to-end counterparts. Ultimately, this may result in less computational resources being consumed when using pre-trained contrastive representation models for as basis for new tasks.

Contrastive learning is not a panacea for all kinds of problems in data modeling and data classification, prediction and clustering, but for a reasonable subset of application types, on certain types of datasets it is a suitable approach to improve performance on downstream tasks. Nor is it an approach with all of its problems and issues solved, and in the paper we highlighted areas for future research, some of which are fundamental issues.

For practitioners who want to apply contrastive methods for pre-training representations on different datasets, we suggest to be mindful about:

- Any inherent characteristics and biases in the data set, e.g. do the images contain only one or multiple objects, are the objects in the center, etc.
- The desired properties of the representation for downstream tasks, e.g. occlusion-invariance, color-invariance, temporal-covariance, etc.
- The ways positive and negative pairs are constructed, such that they provide good learning signals and convey the desired properties.

### ACKNOWLEDGMENT

The authors would like to thank Ting Chen from Google Brain for the helpful comments on an early draft of this paper.

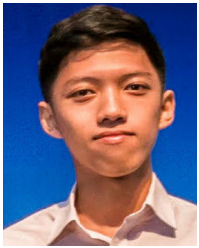

**PHUC H. LE-KHAC** was born in Gia Lai, Vietnam. He received the B.Sc. degree in computer science from Vietnamese-German University, Vietnam, in 2018. He is currently pursuing the Ph.D. degree with ML-Labs, Dublin City University.

Since 2018, he has been a Deep Learning Engineer. His major interest is in representation learning, with a focus on learning representation from video from uncurated sources without human annotations.

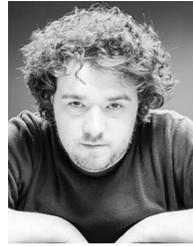

**GRAHAM HEALY** received the B.Sc. (Hons.) degree in computer applications, in 2008, and the Ph.D. degree in brain–computer interfaces, in 2012. He worked as a Postdoctoral Researcher with The University of British Columbia, from 2012 to 2013, and at The Insight Centre for Data Analytics, Dublin City University, in 2013, where he later became a Research Fellow, in 2017, and then became an Assistant Professor in Computing, in 2019. He is currently an Assistant Professor with the School of Computing, Dublin City University. He is interested in the ways computerized systems can automatically detect things from people using signals, i.e., bioelectric, social, collaborative, etc., and then do something useful with that information. His research is a mix of basic-research with a practical focus on developing real-world applications.

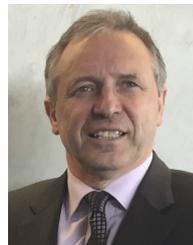

**ALAN F. SMEATON** (Fellow, IEEE) was born in Dublin, Ireland. He received the B.Sc., M.Sc., and Ph.D. degrees in computer science from University College Dublin, in 1980, 1982, and 1987, respectively.

Since 1987, he has been on the Faculty at Dublin City University, where he has previously served as the Head of the School of Computing and the Dean of Faculty. He is a Founding Director of the Insight Centre for Data Analytics, one of the largest publicly-funded research centers in Europe, and was appointed as the Professor of Computing in 1997. He is the author of more than 600 research papers and book chapters with more than 17 700 citations, and has an h-index of 67. His major research interest is in helping finding people to find information, and trying to discover why they need that information and if it is information they previously had, why they have forgotten it.

Prof. Smeaton is an elected member of the Royal Irish Academy and the Winner of the Academy's Gold Medal in Engineering Sciences, in 2016. He is a member of the ACM and the Current Chair of the ACM SIGMM (Special Interest Group in Multimedia).

∙ ∙ ∙